# Resolution of Difficult Pronouns Using the ROSS Method

Glenn R. Hofford, Software Engineering Concepts, Inc.

## Abstract:

A new natural language understanding method for disambiguation of *difficult* pronouns is described. Difficult pronouns are those pronouns for which a level of world or domain knowledge is needed in order to perform anaphoral or other types of resolution. Resolution of difficult pronouns may in some cases require a prior step involving the application of inference to a situation that is represented by the natural language text. A *general method* is described: it performs entity resolution and pronoun resolution. An *extension* to the general pronoun resolution method performs inference as an *embedded commonsense reasoning method*. The general method and the embedded method utilize features of the ROSS representational scheme; in particular the methods use ROSS ontology classes and the ROSS situation model.

ROSS ontology classes include the *object frame class* and the *behavior class.* The ROSS behavior class defines associations among a set of objects that have attribute-based state descriptions and nested behaviors. In addition to the classes of the ontology, the methods use several working memory data structures, including a *spanning information* data structure and a *pronoun feature set* structure. The ROSS internal *situation model* (or "*instance model*") is an instance of a meaning representation; it is a spatial/temporal representation of declarative information from the input natural language text.

A new representational formalism called "*semantic normal form*" (SNF) is also introduced. This is a specification at the abstract level for a set of data structures that are used to store the syntax and content of input natural language text that has been transformed and augmented with semantic role and other information. It is an intermediate form of the input information that is processable by a semantic NLU engine that implements the pronoun resolution method.

The overall method is a working solution that solves the following Winograd schemas: a) trophy and suitcase, b) person lifts person, c) person pays detective, and d) councilmen and demonstrators.

Many of the features described in this paper have been productized - the functionality is implemented in an NLU system that is available for use via a RESTful API server *(currently English-only).*

***Contact:*** *glennhofford(at)gmail.com*

## Table of Contents

## 1. Introduction and Background

Disambiguation of so-called “difficult” pronouns is a challenging problem for natural language processing. Although statistics-based approaches are at least partly effective for some cases, the problem calls for a semantic approach that addresses the representational aspects using deeper and more powerful techniques that involve comprehension of the meaning of natural language. The application of world knowledge and domain knowledge seem to be essential components of the cognitive processes that are used by us as humans in order to comprehend language, i.e. to grasp its meaning in a manner that allows for a reasoning process that reaches conclusions regarding the meaning (i.e. the referent) of pronouns in natural language text or spoken discourse. The challenge lies in somehow emulating this approach in software.

A new general-use ontology-based artificial intelligence method is presented that uses a complex multi-stage set of algorithmic processes that effectively resolves important categories of ambiguous pronouns. The method creates an internal *situation model* of the subject matter of natural language text that enables identification of the referent and the antecedent of a pronoun[1]. A ROSS situation model is an internal memory representation of instances of objects and processes that is constructed as part of the natural language understanding process. The method uses an ontology-based approach that involves ROSS *object frame classes* and *behavior classes*. The classes of the ontology are directly involved in the creation of the situation model as they provide a basis for the instantiation of object instances and process instances.

An important extension to the basic method is also described. This extension involves an *embedded inference process* that performs commonsense reasoning and that is invoked for pronoun resolution problems that are not adequately handled by the basic resolution method. The embedded inference routine specifically handles natural language sentences wherein there is an indirect association between the semantics for the unresolved pronoun and the set of candidate referents. (A specific example is presented from Winograd schema #1 (“councilmen and demonstrators”).

A second extension is described as it applies to a solution for Winograd schema #2 (“trophy and suitcase”). With this extension, the basic (general) method can be supplemented by the use of a set of ontology classes and situation model features that model not only the semantics of the natural language text, but also the “meta” entities and aspects of the communication process itself: these include the “communicative agent” (the talker), the information that is communicated, the receiving, or “self” agent (the listener), and cognitive processes on the part of the communicative agent or agents.

Where the ontology is small, the task of difficult pronoun resolution can be addressed without the use of probabilistic representations in the ontology and situation model and without probabilistic reasoning. However, the introduction of probability data into the ontology becomes necessary in order to scale the method. The ROSS representational scheme has support for probability fields for attribute types, attributes, structure and for nested behaviors within behavior classes. The use of the behavior class probability field is demonstrated by the solution to variant #1 of Winograd schema #1 (“councilmen … feared violence”).

[1] *referent* is used here to denote the external thing; *antecedent* denotes the syntactic item (usually a word or phrase).

### 1.1. The ROSS Representational Method

The ROSS method (Hofford 2014 (a, b)) is a new approach in the area of representation that is useful for many artificial intelligence and natural language understanding (NLU) tasks. (ROSS stands for “Representation”, “Ontology”, “Structure’”, “Star” language). ROSS is a physical symbol-based representational scheme. ROSS provides a complex model for the declarative representation of physical structure and for the representation of processes and causality. From the metaphysical perspective, the ROSS view of external reality involves a 4D model, wherein discrete single-time-point unit-sized locations with states are the basis for all objects, processes and aspects that can be modeled.

The ROSS method is also capable of the representation of abstract things – they are modeled by grounding them in a 4D space-time model. Abstract entities that are modeled include the entities that are involved in the representation of representation (“meta-representation”), including representation of intelligent agent mental representations, cognition and communication.

ROSS is used in two ways in support of the pronoun resolution and inference methods: 1) the *Star ontology language* is used for the specification of *object frame classes* and for rule-like constructs referred to as *behavior classes* in the ontology/knowledge base, and 2) the formal scheme of the ROSS *situation model* (also called “*instance model*”) is used for the specification of meaning representations that represent the semantics of a particular situation.

The ontology+knowledge base repository stores supporting definitions, object frame classes, and representations of conceptual, or world knowledge that use the behavior class. The ontology and knowledge base is organized into three tiers: an upper tier contains supporting definitions and high-level abstract classes, a middle tier contains classes whose primary purpose is functional: middle tier classes are used in many behavior classes, and a lower tier of object classes contains a large number of classes that are distinguishable from other similar classes by a few features. Examples of lower tier classes include “house cat”, “trophy”, and “father-person”.

The internal instance model that is used during processing is a proprietary feature of ROSS that is used for representing factual information about particular situations (past, present or hypothetical situations).

### 1.2. Background: Winograd Schema Challenge

The Winograd Schema (WS) Challenge (Davis: 2011a ) is a set of tests for assessing whether or not an automated natural language understanding system has capabilities for "thinking" - i.e. does the system use and exhibit true intelligence in some sense, or is it responding to human-entered natural language input using canned (hard-coded) replies, "tricks", deception, diversion from the topic, etc. The WS challenge includes a variety of schemas: a schema consists of a pair of descriptive sentences and an associated pair of questions that tests whether or not the system has understood the sentence and its alternate. The NLP task involves some form of anaphora or coreference resolution for an ambiguous, or difficult pronoun that exists in the original sentence. The purpose of the WS Challenge is not to test for simple disambiguation; rather it is to use this task as a test of underlying intelligent capabilities.

The fields of commonsense reasoning for AI and NLU and of anaphora resolution and related disambiguation tasks can be explored from many perspectives. Nevertheless the author has focused particularly on the Winograd Schema Challenge based on the belief that this set of schemas provides a broad-based foundation by its inclusion of a wide variety of problem types that form a sort of “core set” of use cases for NLU.

Davis (2011a) describes the Winograd Schema Challenge as follows:

> A Winograd schema is a pair of sentences that differ in only one or two words and that contain an ambiguity that is resolved in opposite ways in the two sentences and requires the use of world knowledge and reasoning for its resolution. The schema takes its name from a well-known example by Terry Winograd (1972)
>
> > *The city councilmen refused the demonstrators a permit because they [feared/advocated] violence.*
>
> If the word is ``feared", then ``they" presumably refers to the city council; if it is ``advocated" then ``they" presumably refers to the demonstrators.

The schema challenge sentences and test questions for the trophy and suitcase example is described in Levesque et al (2012) as follows:

> The trophy doesn’t fit in the brown suitcase because it’s too big. What is too big?
>
> Answer 0: the trophy
> Answer 1: the suitcase

The obvious answer to a human is that it is the trophy that is too big. This answer is obvious to a person at least partly because humans are able to form a picture (a conceptualization) of the situation as it involves physical objects, processes, and causality. A human is also able to do reasoning about the processes and the causality – i.e. the causal relationships – that are involved in such a situation.

Levesque et al (2012) describes a further aspect of the WS schema challenge for this particular schema:

> …

> 4. There is a word (called the *special* word) that appears in the sentence and possibly the question. When it is replaced by another word (called the *alternate* word), everything still makes perfect sense, but the answer changes.
>
> …
>
> This is where the fourth requirement comes in. In the first example, the special word is “big” and its alternate is “small;” and in the second example, the special word is “given” and its alternate is “received.” These alternate words only show up in alternate versions of the two questions:
>
> • The trophy doesn’t fit in the brown suitcase because it’s too small. What is too small?
>
> Answer 0: the trophy
> Answer 1: the suitcase

Levesque et al (2012) shed light on why this challenge is an appropriate one for purposes of testing whether or not a system that purports to do intelligent thinking and natural language comprehension is actually doing such thinking:

> The claim is that doing better than guessing requires subjects to figure out what is going on: for example, a failure to fit is caused by one of the objects being too big and the other being too small, and they determine which is which.

Addressing this topic with another example involving the aforementioned city councilmen and demonstrators as originally conceived by Terry Winograd, they further state:

> This was the whole point of Winograd’s example! You need to have background knowledge that is not expressed in the words of the sentence to be able to sort out what is going on and decide that it is one group that might be fearful and the other group that might be violent. And it is precisely bringing this background knowledge to bear that we informally call *thinking*.

In his commentary on the difficulty of the “councilmen/demonstrators” example, Winograd (1972) states:

> “We understand this because of our sophisticated knowledge of councilmen, demonstrators, and politics – no set of syntactic or semantic rules could interpret this pronoun reference without using knowledge of the world.”

Solutions for the following W.S. schemas are presented here: schema #1: “the councilmen refused the demonstrators a permit”, #2: “trophy that doesn’t fit in a suitcase”, #8: “the man could not lift his son”, and #115: “Joe paid the detective”. The present method uses a mixture of techniques in solving these schemas – this method is not a “one size fits all” approach.

It will be shown that some disambiguation tasks can be adequately handled by an approach that relies on characteristics that are unique to common objects based on a determination of the higher classes from which they can be said to derive functional properties (i.e. a suitcase is a member of a container class that can be “fitted into”). Other tasks involve a prior-stage determination of semantic roles (active or passive) due to the fact that multiple objects of the same class are involved (“the man could not lift his son” – a person cannot lift another person). Some resolution problems require knowledge that associates behaviors with what are referred to as “nested behaviors” (or “chained behaviors”). The schema that contains “Joe paid the detective …” requires this approach. Finally, many pronoun resolution tasks require the application of a set of preliminary inferences (commonsense reasoning), using a generate-and-test approach that uses temporary situation representations of the situation that is described in order to test candidate antecedents for the pronoun. (This is demonstrated for the “advocate violence” variant of the “city councilmen and demonstrators” schema).

The probability aspects are also addressed with the councilmen and demonstrators schema: it will be shown that the present method can handle this type of resolution problem using behavior classes, or rules, that incorporate a probability value. The example involves an examination of multiple behavior class rules that represent the act of “refusing something with fear as a causal feature”, where the causal connection of this behavior to a prior (nested) behavior in one of the rules is compared with that of other (possibly multiple) rules.

## 2. Main Concepts

### 2.1. Entity Resolution Using a ROSS Ontology

A ROSS ontology is a repository that contains declarative information about objects and processes. Pronoun resolution involves a preliminary stage task that identifies, or links, antecedent words or phrases with items in the ROSS-based ontology. This is referred to herein either as “entity resolution” or as “class selection”. (There is possible overlap with *word-sense disambiguation* which is viewed as describing a closely-related task that is not directly relevant to this method).

Note that there is no single authoritative ROSS ontology; ROSS ontologies are interchangeable. However a single ontology does exist that supports the pronoun resolution examples described in this document.

## 2.2. Ontology Scalability

To support scalability, the ontology that supports the procedures of the resolution method must be general purpose as a declarative representation of entities and features for a problem domain (in this case the commonsense reasoning domain). The ontology should not contain entities, attributes or features that are custom-designed for specific procedural pronoun resolution problems. This may at first appear to be the case for the "trophy and suitcase" schema solution, however it will be shown that the ontology features for that particular schema are generally-useful.

The rationale for the requirement of generally-useful ontology classes and attribute types is scalability: the method can only scale if it depends solely on a set of ontology definitions that have been created or derived apart from considerations of problem specificity.

## 2.3. The ROSS Instance Model

The ROSS instance model has an important role in supporting the pronoun resolution and inference processes. Declarative content of the input natural language text is used in order to build a central instance model that contains a semantic representation of all objects and processes that can be identified during the execution of the entity resolution processing task. *(Note that a situation model is a type of instance model; the terms are used interchangeably in this document).* The instance model thus contains a set of referents (referents can be either objects or processes, however for the examples of this paper they are objects). The information in the instance model is also tracked by an internal memory data structure called the "spanning information stack". Spanning information is tied into the instance model and is used for tracking referents with respect to their level of immediacy to the phrase or clause that contains the unresolved pronoun. The task of pronoun resolution can thus be re-stated as a task that involves a determination of which instance-model-based referent is indicated.

The instance model is not limited to containing objects or processes that are explicit in the input NL text: for instance exophoric pronouns refer to objects that are not explicitly described but that can be represented in an instance model. An example of a sentence with an exophor is "Nobody came to the beach party because it was too hot". Although this case may perhaps be interpreted in any of several ways[2], it can be adequately addressed using a ROSS behavior class that represents weather phenomena (in the locality of the beach where the party would be held) via a behavior class specification that represents a collection of air molecules. As a commonsense representational problem – not a physics problem –an attribute type such as "RelativeTemperatureExperiencedByPersons" may be adequate as an abstraction for representing the state of being "too hot".

---

[2] The pronoun "it", within "it was too hot" may be viewed either as an exophoric reference or as a pleonastic pronoun.

## 2.4. Features of the ROSS Behavior Class That Support the Resolution Process

ROSS behavior classes have a prominent role in providing a set of referent target options; these include objects that are represented by common nouns and nested behaviors that can be represented by either nouns or verbs.

The behavior class has the following features that support the resolution and inference processes:

- Multiple time and space-related constituent elements within a single behavior class, where elements can be:
  - Physical objects: what is actually stored is the state or states of an object (as specified using ROSS attributes); the physical object and its specified state are part of a wrapper class called a "populated object class".
  - Nested behaviors (e.g. one of possibly many behavior classes for "refusing a permit request" can contain a nested behavior class that represents "fearing a harmful event").
- Support for the representation of un-communicated objects (see the "beach party" example above). Behavior classes can involve a wide variety of objects and nested behaviors that are implicit in a situation: in addition to phenomena like the weather, these may include the ground (earth) and persons that are observers.
- A representational construct called the "binder" that allows for the representation of the spatial and temporal relationships between the various objects that are part of a behavior.

*(Hofford (2014 (b)) "The ROSS User's Guide and Reference Manual" describes the behavior class in greater detail).*

## 2.5. Definitions

The following terms are unique to the present method or have unique uses pertaining to the method.

- **meaning unit**: a meaning unit ("ME") is a syntactic construct that consists of a subject, a predicate and any adverbial modifier words, phrases or clauses. The predicate contains verb-based expressions and includes objects (direct object and indirect object). Meaning units are recursive and nested MEs may occur in any of several places. Generally speaking a meaning unit is the equivalent of a clause. There is usually a one-to-one correspondence between a syntactic ME and a semantic predicate expression, described next. Examples of meaning units include: "Bob did walk the dog.", and "because it was too big".
- **predicate expression**: (part of semantic normal form (SNF)) - a (semantic) predicate expression ("PE") is a semantic construct that centers around a single syntactic predicate expression (e.g. "*did walk the dog*"). PEs have arguments that have roles such as "actor" and "actee". Like MEs, PEs can be nested. The PE is explained in greater detail below.

## 2.6. Overview of the Algorithm

The pronoun resolution general algorithm is part of a larger algorithm called the *semantic engine driver*. The pronoun resolution general algorithm is driven by pronoun instances as they are encountered during execution of an *entity resolution* routine that itself is invoked within the control flow of the engine driver. When a pronoun is encountered, an attempt is made to resolve its referent and the antecedent word or syntactic phrase that corresponds to the semantic entity. (The referent/ semantic entity is not limited to "objects" – e.g. it could be a process or a fact).

The semantic engine driver processes a list of semantic normal form[3] predicate expression (PE) data structures that corresponds to one or more input NL sentences[4]. In a *typical* situation that involves an anaphor, the input NL text fragment consists of at least two consecutive meaning units, a *main* meaning unit and a second (*current*) meaning unit that contains one or more unresolved pronouns. Several tasks are applied – the processing described here starts with the main PE, which represents the main meaning unit.

| **Example:** | |
|---|---|
| Main meaning unit: | "The trophy doesn't fit in the brown suitcase" |
| Current meaning unit: | "because it's too big" |

The first task involves *class selection* (entity resolution) for all common nouns and proper nouns in the main PE (and possibly pronouns, based on earlier resolution results). In the example shown this would involve selecting a TrophyClass and a SuitcaseClass. The selected classes are then used for the second main task: *instantiating object instances* within the master internal instance model. For this example this creates a new "trophy" object instance and a "suitcase" object instance within the master instance model.

The third task also involves use of the main PE: it is a form of entity resolution referred to as *behavior class selection*: this selects a behavior class or list of behavior classes that are relevant for the situation. The behavior class selection process takes into account not only the verb word (e.g. "fit", "lifted", "payed", "refused") but whether or not the event or action is negated, and whether or not the active, passive and "extra" object instances are a match with respect to their class, or higher class in an inheritance hierarchy, and with respect to their use in active, passive or extra roles. The method is further capable of utilizing verb modification phrases (usually adverbs or adverbial phrases) in the input (e.g. "completely fit", or "tightly fit"; e.g. "walking quickly" versus "walking", and "trying to walk" versus "walking") – this input guides the behavior class selection process via a process that matches verb modification information against behavior class modification parameters.

[3] See section *4. Semantic Normal Form* for a description of semantic normal form.

[4] Predicate expressions (PEs) and meaning units (MU) are used somewhat interchangeably throughout this document. The actual method as it has been implemented involves a semantic engine that uses MEs as input; however, the description of the method will usually utilize the PE as the basic input building block.

Fourth task: once the list of behavior classes, each of which matches all search criteria, has been attained, the NLU system is able to fully describe the situation of the main meaning unit ("The trophy doesn't fit in the brown suitcase"). (Note that each of the retrieved behavior classes in the list are equivalent with respect to the information that they provide for instance model generation). The first behavior class in the list is used to generate new object instances in the master instance model. *(An alternate approach is to use a higher behavior class rather than the first of multiple similar behavior classes).* The step of generating new object instances using the behavior class is called "behavior class application". (Details of this process are outside the scope of this document).

The transition to the next task involves completion of processing of the *main* PE and the start of the processing of the *current* PE.

The fifth task involves processing of the entity arguments of the *current* PE: this starts with entity resolution and instance model generation for *any* entity arguments that do not contain pronouns. (The trophy and suitcase example does not have any such entity arguments in the current PE). (Where the current PE only contains adjectival information (as in "too big"), this will get saved in the pronoun feature set data structure).

The sixth task involves processing of the entity arguments of the current PE that contain pronoun(s). (For the trophy and suitcase example, this involves processing of the entity argument containing the "it" of the current PE). An early part of this process is entity resolution: it in turn involves the actual pronoun resolution. The pronoun resolution routine involves a search for the constituent element – usually an object instance - of the master instance model that matches the features of the unresolved pronoun, as they are specified or implied in the current meaning unit (and current PE). The search process is limited to those instance model object instances that are associated via pointers from a *spanning information* data structure. The search process involves examination of each of the following to find a match (note that all criteria that are provided by the text of the current meaning unit (represented in the *pronoun feature set*) are necessary for a match to succeed).

- The ***pronoun feature set:*** all features of the unidentified object or event that is represented by the unresolved pronoun: this includes all of the following that exist. *(Note: the features here are described using various possible trophy and suitcase sentences).*

  - an associated attribute or state if one exists (e.g. "because *it* is *too big*."). *Matching against instance model:* match this feature against an optional causal feature attribute for a populated object class within the behavior class that is associated with an object instance.
  - a behavior of the meaning unit in which the pronoun is contained (e.g. "because the packing person did not *push it* hard enough") (*it* participates in a *push* behavior). *Matching against instance model:* match this feature against a nested behavior in the behavior class.

- the active/passive/extra role within the meaning unit. (e.g. "because *it* was not pushed hard enough" (*it* has passive role). *Matching against instance model:* match this feature against a *PassiveParticipant* flag that belongs to passive role populated object classes within a behavior class.

- ***Instance model features***: qualitative attributes/states, spatial/temporal relationship to other objects within the instance model, object frame class or higher class in the hierarchy, and active/passive role. These features may be determined by information in the instance model itself, or indirectly via an inspection of the behavior class that was used in generating the instance model objects from the main meaning unit. If the behavior class is involved, the populated object classes or nested behaviors are examined. Note that the instance model and spanning information structure may in some cases include object instances that are not explicit in the text: this is possible where a behavior class was applied to the main meaning unit and resulted in the generation of non-explicit object instances (e.g. the weather, e.g. the ground). In such cases, an exophoric pronoun will be matched against the object instance.

If the pronoun referent and corresponding syntactic antecedent can be resolved, both the instance model object instance and its class are associated with the pronoun, and this newly-acquired information is added to the instance model. (Subsequent processing may also use the newly-acquired semantic information (pronoun class and object instance) during application of a behavior class for the current PE). The new information that identifies the pronoun is also added to the spanning information data structure for possible subsequent use. If the pronoun is not resolved via the matching process described above, other resolution attempts can be made: these include matching based on gender or number. Finally, a default resolution mechanism is invoked if all other resolution attempts have failed; in this case a return code indicates that pronoun resolution did not succeed using the instance model-based approach: this allows for subsequent processing to handle possibly-cataphoric pronouns.

The spanning information data structure keeps track of classes and instances for each *main* meaning unit/PE so that a *current* meaning unit/PE may refer to them. The *spanning information stack* extends this concept by keeping track of the classes and instances for up to *n* prior meaning units, where the value of *n* is chosen based on practical considerations.

### 2.7. Probability-Based Pronoun Resolution

The functionality of the method has been described apart from the use of probabilistic information that may be available. Both the entity resolution method and the pronoun resolution method can be supplemented by using probability fields within the classes. The probabilistic functionality for pronoun resolution will be explained and demonstrated as it has been applied to the "feared violence" variant of the councilmen and demonstrators schema.

### 2.8. Use of ROSS Situation Model to Support Question Answering

Once a situation/instance model is generated by the semantic engine, it can be used for a variety of follow-up tasks; a primary example is that of question answering. For instance, for the "man lifting son" schema, the follow up question "Who was so weak?" is processed by searching the instance model that was previously generated when the original sentence was processed.

### 2.9. Optional Representation of the Communicative Agent

The method can also incorporate an *optional* model that represents intelligent/communicating agents, information that is communicated, and cognitive information and processes. (See *Appendix 1: Solution for "Trophy and Suitcase" Schema Using a Model of the Communicating Agent* for full details). This optional approach involves generation of extra "meta" information in the instance model so that the reception of natural language input is represented as a process that involves one or more communicative agents (a "talker" or "talkers"). The communicated information is also represented in the instance model. The information is received by a *self-agent* (the "listener"), i.e. the NLU system, which can also be represented in the instance model.

The general pronoun resolution method described in this document does not include considerations of modeling of the communicative agent and cognition. It makes a set of default epistemological assumptions: that there is a shared ontology, and that the communicative agent adheres to a set of shared rules (e.g. about causality in the physical world) in the realm of cognition; this allows the tasks of entity resolution and pronoun resolution to be handled using an approach that deals directly with the input text and the semantics of the text.

### 2.10. Non-Objective: Representation of Deep Structure of Physical Objects

Some lines of research in the area of commonsense reasoning have focused on spatial representations and spatial reasoning. This approach is exemplified by Davis (2011b), wherein he describes the trophy and suitcase example. He states

"The first task is to interpret the phrase, "because it was too large" in terms of its spatial content."

In his subsequent analysis, he emphasizes the spatial reasoning aspects of the problem.

The present method takes a different tack: it relies on class inheritance that involves a *middle ontology* that includes classes such as "container", or "two-sided enclosure", and "enclosable object". These middle ontology classes have attribute types such as "size relative to the process of fitting", from which can be derived attributes with values such as "too big" or "too small". Lower ontology objects like trophies and suitcases derive some of their features from the higher classes (e.g. container) that they are associated with via the inheritance mechanism. The anaphora resolution method is focused on the task of identifying the entity that is the most likely of the candidate referents.

The present approach does not fully emulate human thought processes as they are used to disambiguate pronouns; in some respects it is based only on useful abstractions. As such it does not handle all conceivable pronoun resolution cases: a method that employs a deep structure representation of the objects of a situation may indeed be necessary for many such cases. Such a method could be based on ROSS, and would represent the following aspects:

- Instantiation of object instances using values that represent compositional properties, e.g. "substance" properties. For instance, this approach involves representations of common objects like trophies and suitcases with respect to whether each unit-sized cubicle region (e.g. with dimensions the size of a millimeter) is solid or space. Further depth of analysis and representation involves questions regarding aspects such as flexibility of materials (or the lack thereof) (e.g. a cloth suitcase may be flexible in various parts thus allowing something that seems too big to actually fit into it).
- Specifications of the sizes of objects and of all distances between objects.
- The spatial orientation of all object instances.
- Behavior classes (causal rules) that redefine coarse-grained rules such as "fitting" in terms of fine-grained rules such as a rule that describes that a solid-filled cuboid region at t=1 cannot occupy the same position as another (adjacent) solid-filled cuboid region at t=2 unless the other solid has "moved out of the way".

The author's view is that the ROSS method is a promising approach for achieving the deep spatial reasoning that would accomplish anaphora resolution using the above guidelines.

### 2.11. Comprehendor NLU System

Comprehendor is a natural language understanding (NLU) system that performs a variety of NLU tasks. While this paper describes a method and a main set of use cases for difficult pronoun resolution, it also describes a supporting set of uses cases for ontology derivation and knowledge acquisition as performed by the Comprehendor system. The ontology derivation/knowledge acquisition capabilities are viewed as significant in their own right; they have provided a substantial boost in time-savings for purposes of tackling new disambiguation method use cases. (The ontology derivation and knowledge acquisition sub-system is a separate topic of research and development by the author as part of an ongoing effort to create a controlled natural language for ROSS).

## 3. Pronouns: Types of Pronouns and Syntactic Locations of Pronouns

### 3.1. Types of Pronouns Handled by the Method

The class of difficult pronouns that is handled by the method includes the following types of pronouns:

- Personal subjective: he, she, it, they
- Personal objective: him, her, it, them

First and second person personal pronouns require somewhat different handling and are viewed by the author as a part of the area of modeling the intelligent agent (not included in this document). Other classes of pronouns include possessive, demonstrative, Wh-pronouns, reflexive and interrogative: resolution of some of these categories of pronouns does not yield to the present method.

### 3.2. Pronoun Syntactic Locations

The resolution of the third-person personal pronouns that are the focus of the present method involves a process of analysis that centers on, (or "pivots" around) the imaginary dividing line between a *pair* of *adjacent* meaning units. Other configurations are handled as secondary cases – these include antecedents that are several clauses or sentences back, and exophoric pronouns. The following are the primary configurations for personal pronouns and their antecedents as they appear within the syntactic structure of natural language sentences:

- Anaphora crossing meaning units: the pronoun is within a *current* meaning unit and the antecedent is in an earlier meaning unit. Variations include but are not limited to:
  - A main clause (earlier) containing the antecedent, followed by an adverbial clause (current) that contains the pronoun as a (noun phrase) subject. (e.g. "The man could not lift his son because he was too weak.").
  - A main clause (earlier) with antecedent, followed by an adverbial clause (current) that contains the pronoun as direct object, or that contains the pronoun within a prepositional phrase complement. (e.g. "The man could not lift his son because the building had collapsed on top of him.").
- Cataphora crossing meaning units: the pronoun is within a *current* meaning unit and the antecedent is in a later meaning unit. (e.g. "When he arrived home, John went to bed.").
- Anaphora within a meaning unit: the current meaning unit contains a sentence with a personal objective pronoun that refers to an antecedent within the same meaning unit. This structure is shown by these sentences: "The house's owners sold it last year.", or "The owners of the house sold it.".

## 4. Semantic Normal Form (SNF)

This section contains a formal specification of the input needed by a semantic engine that implements the present method; this is referred to as *semantic normal form* (SNF)[5]. Semantic normal form is a syntax-independent formalization; it is an intermediate representation that stands between syntax and the ROSS instance model. SNF has been designed to facilitate instance model creation.

The data structure definitions here may be used for the creation of engine input data adapters; this allows for flexibility with respect to parsers that can be integrated into systems that use the present method. SNF is language-independent and thus allows use of the present method with a wide variety of natural languages.

### 4.1. The Predicate Expression ("PE")

The *predicate expression* ("PE") is the basic building block of semantic normal form[6]. A predicate expression consists of a *predicate specifier list*, a list of *entity argument specifiers*, or entity arguments, a list of *attributive argument specifiers*, or attributive arguments, and a list of *modification specifiers*, or modifiers. Entity arguments are typically associated with semantic entities that correspond to the syntactic subject, direct object, indirect object, and those that are represented by nouns or noun phrases within post-verb (predicate complement) prepositional phrases. Attributive arguments are words or phrases that represent attributes (usually representing an adjective used with a form of "to be"). Modifiers are associated with adverbial syntactic items, e.g. adverbs and adverbial phrases and clauses. Predicate expressions allow for indirect recursion, or nesting: an argument may itself be a predicate expression, a modifier may be a predicate expression or it may be a modification specifier expression that includes a predicate expression.

### 4.2. Semantic Role Labels

#### 4.2.1. Predicate Specifier Roles

The predicate specifier has a *predicate specifier role label*. This label has one of the following enumerated values (this list is not exhaustive). (Actors/actees/extras are explained in the following section).

```
enumeration PredicateSpecifierRole
{
   PredicateToBeAttributive,            // "The sky is gray."
   PredicateToBeIsA,                    // "A car is a vehicle."
   PredicateHasAVerb,                   // "A vehicle has wheels."
```

[5] Although the term "semantic normal form" may have other prior use(s), the author is unaware of any restrictions regarding its use; any overlap with other concepts represented by the term are unintentional.

[6] The term "predicate expression" connotes that it contains representations that correspond to syntactic expressions; neither the predicate expression itself nor the immediate constituent fields of a predicate expression are themselves true expressions. Note that "predicate unit" is also used as a synonym for "predicate expression".

```
    PredicateToBeTakingEntityArgument  // "The car is in the garage."  (with actor and extra)
    PredicateVerbTakingEntityArgument  // "The man walked."  (with actor)
                                       // "The man lifted his son."  (with actor and actee)
                                       // "The ball was thrown."  (with actee)
}
```

Note that syntactic concepts such as auxiliary/helper verb uses of "to be" are not present here. E.g. for the sentence "The ball was thrown", the predicate specifier verb word is the "throw" verb, the role is PredicateVerbTakingEntityArgument, and "was" is not stored in the data structure.

### 4.2.2. Entity Argument Roles

An entity argument has an *entity argument role label*. This identifies the argument as *actor* (active, or causative role), *actee* (passive role) or *extra* (neither active nor passive role).

```
enumeration EntityArgumentRole
{
    Actor,
    Actee,
    Extra
}
```

Entity argument roles have their syntactic origination in syntax categories such as subject and direct object, however they are a reflection of the need to represent the phenomenon of causality. The determination of an entity argument role may involve a syntactic analysis of a prepositional phrase: e.g. the sentence "The man was bitten *by the dog*" gets processed to generate two arguments: "dog" gets the actor role, and "man" gets the actee role. The extra role is for entities that are neither active nor passive. Extra role entities often come from prepositional phrases that are complements of the main verb or predicate; an example of an entity with the extra role is "building", in "She walked away from the *building*.".

For the "man could not lift his son" schema this results in the following assignments of actor/actee/extra roles:

```
Main Meaning Unit - "The man could not lift his son"

        Actor := man, derived from the subject phrase
        Actee := son, derived from the direct object noun phrase
        Extra := (none)

Subsequent Meaning Unit – "because he was so weak."

        Actor := he, derived from the subject phrase
        Actee := (none)
        Extra := (none)
```

### 4.2.3. Extra Sub-Roles

Extra sub-roles are used for entities that have an association with the semantics of a predicate that is neither active nor passive. Many of the sub-roles are directly derived from prepositions. E.g. for the sentence "The man drove the car around the block.", the word "block" has the *extra* role and the *Around* sub-role.

```
enumeration ExtraSubRole
{
   IndirectObject,  // "the councilmen refused the demonstrators a permit"
   About,
   Above,
   Around,
   At,
   Before,
   From,
   Into,
   Over,
   Under,
   // (others here not shown)
}
```

## 4.3. Relative/Subordinate Clauses

Relative/subordinate clauses do not supply entities to the predicate expression in which they are contained; rather, they consist of: a) a possible preposition, e.g. "from", b) Wh-pronoun, e.g. "who", and c) a nested predicate expression that has its own set of entity arguments. Examples include "The sheriff arrested the man *who had held up the bank*.", "They followed the stream through the woods to the spring *from which it had its source*." The nested predicate expression is handled differently from entity arguments that do not contain nested PEs: it is processed by an indirect recursive call as a PE that is part of the overall syntactic sequence of PEs (cf. PredicateExpressionPointerList in section *6.Semantic Engine Driver: Data Structures and Control Flow*).

## 4.4. Attributive Argument Roles

Predicate expressions with predicates having the PredicateToBeAttributive role or the PredicateToBeIsA role have attributive arguments. Examples include "The sky is blue", "Mary is seven years old", and "A car is a vehicle".

```
enumeration AttributiveArgumentRole
{
   Attribute,
   HigherClass
   // (others here not shown)
}
```

Examples from the Winograd schemas include “it was too big”, and “he was so weak”.

### 4.5. Other Argument Categories: Abstractions That Represent Aspects

*(This section draft/under review)* This is a list of categories of arguments that do not get handled in the same way as other arguments since they are not instantiated within instance models as objects or as behaviors:

- Aspect types: e.g. color (example: “The intense color of the sky dazzled the observers.”)
- Aspects: e.g. “blueness” (example: “The blueness of the sky extended to the horizon.”)

One option for handling such abstractions is reification of the aspect type or aspect in the ontology and in instance models.

### 4.6. Other Enumerated Types

The *SyntacticRole* is used to represent the syntactic origin (currently limited to use for noun phrases)

```
enumeration SyntacticRole
{
  Subject,
  DirectObject,
  IndirectObject,
  Other
}
```

The *DiscourseContext* enumerated type represents mood+tense.

```
enumeration DiscourseContext
{
  DeclarativePastSimple,
  DeclarativePastPerfect,
  DeclarativePastProgressive,
  DeclarativePastPerfectProgressive,

  DeclarativePresentSimple,
  DeclarativePresentPerfect,
  DeclarativePresentProgressive,
  DeclarativePresentPerfectProgressive,

  DeclarativeFutureSimple,
  DeclarativeFuturePerfect,
  DeclarativeFutureProgressive,
  DeclarativeFuturePerfectProgressive,

  InterrogativePastSimple,
  InterrogativePastPerfect,
```

```
   InterrogativePastProgressive,
   InterrogativePastPerfectProgressive,

   Imperative,

   Hypothetical  // e.g. "if an object is dropped then it will fall"
}
```

### 4.7. The Structure of the Predicate Expression

The structure of the predicate expression is described here using a hybrid form that mixes data structure pseudo-code with BNF. Lower level items are described after the larger items in which they are contained. Optional items are bracketed with '[' and ']'. The order of items within a structure is not important unless specifically indicated or implicit within a BNF expression. A list may contain 0, 1 or multiple items unless otherwise noted. (Note: "predicate unit" is sometimes used as a synonym for "predicate expression"). *(This is a high-level view of SNF and many lower-level items such as PrepositionalPhraseComplement are not defined in detail).*

```
PredicateExpression
{
   PredicateSpecifierList
   EntityArgumentSpecifierList
   AttributiveArgumentSpecifierList
   ModificationSpecifierList
   [ IntroductoryWord ]  // e.g. "that"
}

//========================================
// PredicateSpecifier
//
PredicateSpecifierList  ->  PredicateSpecifier
                            | PredicateSpecifier  PredicateSpecifierList ;

PredicateSpecifier
{
   Ordinal  // 0-based position within the list
   MainVerbWord  // e.g. "likes", "caused", "walk", "running"
   PredicateSpecifierRole  // e.g.  PredicateToBeAttributive
   DiscourseContext
   [TrailingConnectiveWord ] // e.g. "and"
}

IntroductoryWord  ->  "that" | … ;

//========================================
// EntityArgumentSpecifier
//
EntityArgumentSpecifierList  ->  EntityArgumentSpecifier
                                 | EntityArgumentSpecifier  EntityArgumentSpecifierList ;
```

```
EntityArgumentSpecifier
{
   // At least one must exist:
   [ EntityDesignatorList ]  // must be non-empty
   [ PredicateExpression ]
   //
   EntityArgumentSemanticRole   // one of: Actor, Actee, Extra
   ExtraSubRole  // (can be NULL) e.g. Around, Into, Over
   SyntacticRole  // (syntactic origin) - Subject, DirectObject etc.
   PredicateOrdinal  // refers to a predicate specifier
}

EntityDesignatorList  ->  EntityDesignator
                          |  EntityDesignator  EntityDesignatorList ;

EntityDesignator
{
   // At least one must exist:
   [ NounPhrase ]
   [ PrepositionalPhraseComplement ]
   //
   [TrailingConnectiveWord ]  // e.g. "and"
}

NounPhrase  ->  NounHeadWordList
              | [SpecifierList] [QualifierList] NounHeadWord [PostnominalModifierList]
              | NounPhrase BoundRelativeClause ;

NounHeadWordList  ->  NounHeadWord
                      |   NounHeadWord NounHeadWordList ;

NounHeadWord  ->  Pronoun
                  |  CommonNoun
                  |  ProperNounPhrase ;    // e.g. "Earnest W. Quality"

SpecifierList  ->  Specifier   // e.g. "the", "this", "first"
               |  Specifier  SpecifierList ;

QualifierList  ->   Qualifier  // e.g. "angry", "old", "green"
                |  Qualifier  QualifierList ;

PostnominalModifierList  ->  PostnominalModifier  // e.g. "in the garage"
                          |  PostnominalModifier  PostnominalModifierList ;

PostnominalModifier ->  PrepositionalPhrase
                     |  AdjectivePhrase ;

// Note: each of the following may contain nested PEs:

PrepositionalPhrase ->   // (not shown) e.g. "from which its name is derived"

BoundRelativeClause  ->   //  (not shown) e.g. "the man who drives the bus"
```

```
PrepositionalPhraseComplement  ->  //  (not shown) e.g. “in the brown suitcase”

//=========================================
// AttributiveArgumentSpecifier
//
AttributiveArgumentSpecifierList  ->  AttributiveArgumentSpecifier
                                   |  AttributiveArgumentSpecifier  AttributiveArgumentSpecifierList ;

AttributiveArgumentSpecifier
{
   [ AttributeDesignatorList ]
   // (higher classes not shown)
}

AttributeDesignatorList  ->  AttributeDesignator
                          |  AttributeDesignator  AttributeDesignatorList ;

AttributeDesignator
{
   AttributeDesignator
   [TrailingConnectiveWord ]  // e.g. “and”
}

//=========================================
// ModificationSpecifier
//
ModificationSpecifierList  ->  ModificationSpecifier
                            | ModificationSpecifier  ModificationSpecifierList ;

ModificationSpecifier
{
   // At least one must exist:
   [ AdverbialPhrase ]
   [ AdverbialExpression ]
   [ PredicateExpression ]    // e.g. “slipping past the guard”
   //
   SyntacticPosition   // e.g. Leading, PreVerb, InVerbSequence, PostVerb, Final
   PredicateOrdinal  // refers to a predicate specifier
}

AdverbialPhrase  ->  AdverbWord | AdverbPhrase ;

AdverbWord  ->  literal ;  // e.g. “quickly”

AdverbPhrase  ->  …   // e.g. “early in the morning”

AdverbialExpression  // e.g. “while it was still dark”, “when it is not raining”
{
   Wh-Word | AdverbPhraseIntroductoryWord
   PredicateExpression
}

Wh-Word  ->  “while” | “when” | … ;
```

```
AdverbPhraseIntroductoryWord  ->  “because” | … ;
```

### 4.8. SNF Example

Semantic normal form can be illustrated by a predicate expression that represents the following sentence: “The city councilmen refused the demonstrators a permit because they feared violence.”

```
PredicateExpression
{
   PredicateSpecifierList
      PredicateSpecifier
      {
         MainVerbWord  (“refused”)
         MainVerbSemanticRole (PredicateVerbTakingEntityArgument)
         DiscourseContext (DeclarativePastSimple)
      }

   EntityArgumentSpecifierList (
      EntityArgumentSpecifier   // “the city councilmen”
      {
         EntityDesignatorList (
            EntityDesignator
            {
               NounPhrase
               {
                  SpecifierList
                     Specifier (“the”)
                  QualifierList
                     Qualifier (“city”)
                  NounHeadWord (“councilmen”)
               }
            }
          );
         EntityArgumentSemanticRole (Actor)
         PredicateOrdinal (0)   // refers to “refused”
      }
      EntityArgumentSpecifier   // “the demonstrators”
      {
         EntityDesignatorList (
            EntityDesignator
            {
               NounPhrase // (detail not shown)
            }
          );
         EntityArgumentSemanticRole (Actee)
         PredicateOrdinal (0)
      }
      EntityArgumentSpecifier   // “a permit”
      {
         EntityDesignatorList (
            EntityDesignator
            {
               NounPhrase // (detail not shown)
```

```
            }
         );
         EntityArgumentSemanticRole (Extra)
         PredicateOrdinal (0)
      }

   );  // EntityArgumentSpecifierList

   ModificationSpecifierList
      ModificationSpecifier    // “because they feared violence”
      {
         AdverbialExpression
         {
            AdverbPhraseIntroductoryWord (“because”)

            // Nested PE:

            PredicateExpression
            {
              PredicateSpecifierList
                 PredicateSpecifier
                 {
                    MainVerbWord  (“feared”)
                    MainVerbSemanticRole (PredicateVerbTakingEntityArgument)
                    DiscourseContext (DeclarativePastSimple)
                 }

              EntityArgumentSpecifierList (
                 EntityArgumentSpecifier  // “they” (actor role)
                 {
                    EntityDesignatorList (
                        EntityDesignator
                        {
                           NounPhrase
                           {
                              NounHeadWord (“they”)
                           }
                        }
                    );
                    EntityArgumentSemanticRole (Actor)
                    PredicateOrdinal (0)   // refers to “feared”
                 }
                 EntityArgumentSpecifier  // “violence” (actee role)
                 {
                    // (not shown)
                 }
              );  // EntityArgumentSpecifierList

            } // PredicateExpression

         } // AdverbialExpression

         SyntacticPosition (Final)
         PredicateOrdinal (0)  // refers to “feared”

      } // ModificationSpecifier

}  // PredicateExpression
```

## 5. NLU System Architecture and Data Flow

**Figure 1** shows the high-level architecture/dataflow diagram for an NLU system that implements the present method. The diagram is included as background that shows the context wherein the anaphora resolution method operates.

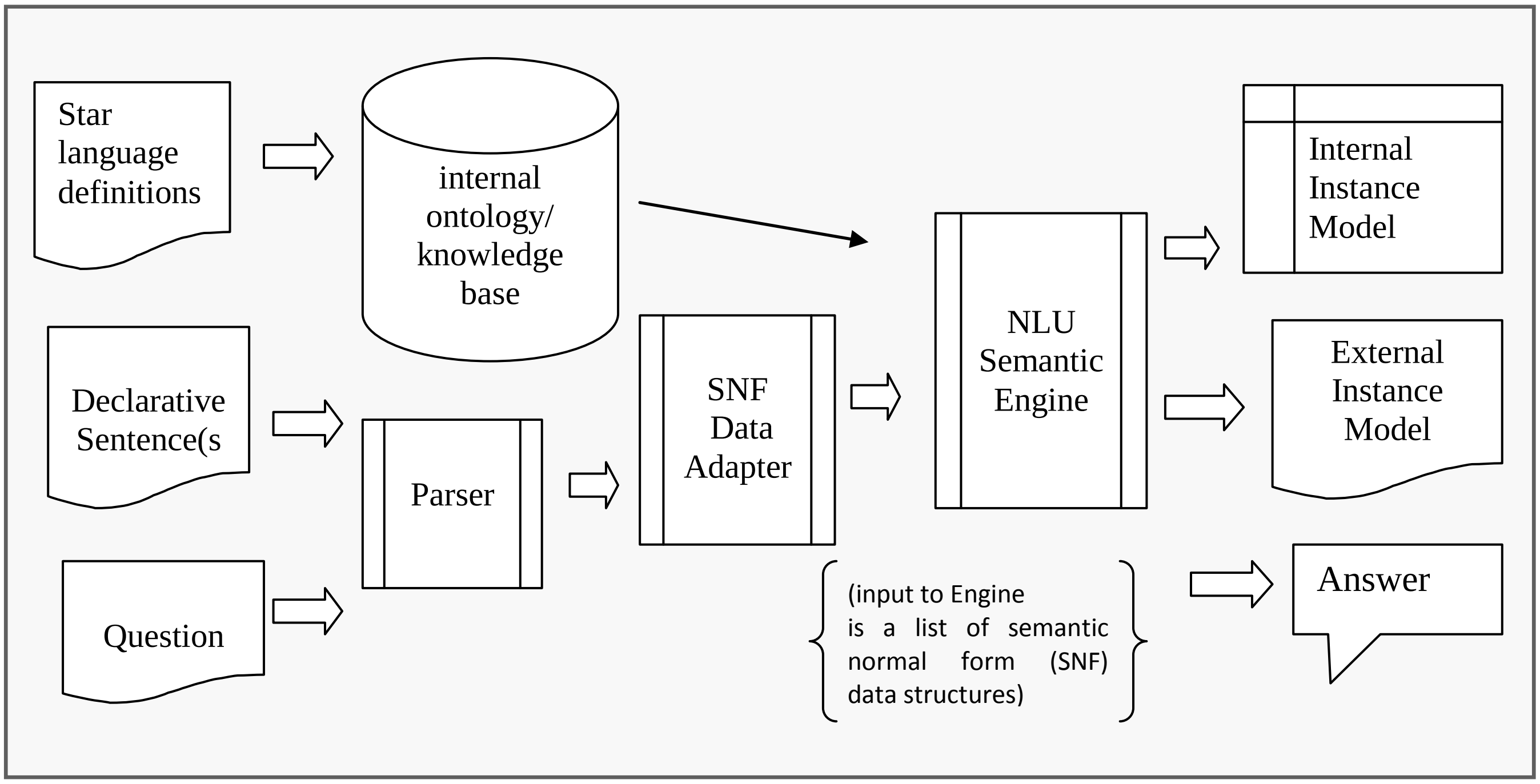

*Figure 1: High-level Architecture of Comprehendor*

A parser subsystem will include a number of subsystems that include lexical analysis, sentence segmentation, morphological analysis, part of speech tagging (possibly optional depending on the parser capabilities), and a parsing component. The parser subsystem generates a list of syntax trees (or syntactic tree-like data structures) that are processed by a SNF data adapter to create a list of SNF predicate expressions (PEs) that are usable by the engine.

The NLU semantic engine subsystem processes the list SNF predicate expressions in order to create an internal instance model. The engine performs the various tasks that accomplish the entity resolution (class selection) and pronoun resolution/disambiguation. The engine uses the internal instance model both for pronoun resolution and for cases where it performs the embedded commonsense reasoning.

### 5.1. Input to a Parser: Communication Unit List

This section describes the structure of the natural language text input in its original form, prior to conversion to semantic normal form.

The root element is Document. A Document is defined as a communication unit list. A communication unit may be a sentence or some other non-sentence textual expression. Non-sentence textual expressions are useful for handling strings of text containing non-sentence text, e.g. news article headlines, date and time stamps, email addresses, etc.

```
Document ->
  CommunicationUnitList ;

CommunicationUnitList ->
  CommunicationUnit
 | CommunicationUnit  CommunicationUnitList ;

CommunicationUnit ->
    SingleWordOnLine
   | TwoWordSequenceOnLine  // e.g. "Chapter 1"
   | DateAndTime
   | EmailAddress
   | WebAddress
   | Sentence ;
```

The following are the grammatical elements under *sentence*.

```
Sentence ->
  SemicolonExpressionList FullStop
 | MeaningUnitList FullStop ;

SemicolonExpressionList ->
  SemicolonExpression
 | SemicolonExpression  SemicolonExpressionList ;

PredicateExpressionOrderedList   -> PredicateExpression
                      |  PredicateExpression CoordinatingConjunction PredicateExpressionOrderedList ;

CoordinatingConjunction  ->  'and' | 'or' | 'but' | ... ;

SemicolonExpression ->
  PredicateExpressionList ';' PredicateExpressionList ;

FullStop ->
  '.' | '!' | '?' ;
```

### 5.2. Data Adapters (Syntax-To-Semantic-Normal-Form Converters)

A data adapter that converts syntactic data, usually consisting of a list of parser-generated syntax trees, to semantic normal form is called an SNF data adapter, or SNF converter. This process provides input in a form that is usable by a semantic engine that implements the present method using SNF.

### 5.2.1. Example: Stanford Parser Output / Data Adapter Input

The following syntax tree (context-free phrase structure grammar representation) was generated by the Stanford parser (online demo at http://nlp.stanford.edu:8080/parser/index.jsp). *(This is provided as an example of possible input to an SNF converter).*

```
(ROOT
  (S
    (NP (DT The) (NN trophy))
    (VP (VBZ does) (RB n't)
      (VP (VB fit)
        (PP (IN in)
          (NP (DT the) (JJ brown) (NN suitcase)))
        (SBAR (IN because)
          (S
            (NP (PRP it))
            (VP (VBZ 's)
              (ADJP (RB too) (JJ small)))))))
    (. .)))
```

### 5.2.2. Example: Phrase Structure Parser Output / Data Adapter Input

The Comprehendor NLU system includes an English phrase structure parser sub-system. This system generates a syntax tree like the following for the trophy and suitcase example sentence (the “too big” variant). The grammar for this parser is not shown, however most of the items shown below have descriptive names that convey their meaning.

```
Communication unit type: Sentence

  Sentence contents:  The trophy [doesn't] does not fit in the brown suitcase because [it's] it is too big.

  Syntax tree:

    MeaningUnit
      SubjectPhrase:
        NounPhrase:
          Specifier List:  The
          Head word: trophy
      PredicatePhrase:
        PreVerbAdverb: not
        AuxVerbWord: does
        MainVerbWord: fit
        Prepositional phrase complement:
          PrepositionalPhrase:
            Head word: in
            NounPhrase:
              Specifier List:  the
              Qualifier List:
                AdjectivePhrase:
                  Head word: brown
              Head word: suitcase
        Final adverbial phrase list:
          AdverbPhrase:
```

```
MeaningUnit
  Introductory word: because
  SubjectPhrase:
    NounPhrase:
      Head word: it
  PredicatePhrase:
    AuxVerbWord: is
    PostVerbAdverb: too
    PostVerbAdjectivePhrase:
      AdjectivePhrase:
        Head word: big
```

*(Note: the test results shown in this document use a version of the Comprehendor semantic engine that directly uses this type of input for each of the schemas).*

## 5.3. Semantic Engine

The semantic engine tasks include the following that are particularly relevant for pronoun resolution.

### 5.3.1. Multi-Stage Process: Main Predicate Expression and Current Predicate Expression

A processing task of instantiating object instances will usually take place prior to that of pronoun resolution (exceptions involve sentences that have pleonastic or exophoric pronouns). This task establishes pointers to entity classes and instances within the spanning information data structure. The *main predicate expression* and the *current predicate expression* are defined as follows. (Note that, syntactically, the main predicate expression may appear after the current predicate expression, as is the case where cataphoric pronouns are involved).

**main predicate expression**: the predicate expression that contains semantic information about previously-resolved entities:

- It is described by a *spanning information* data structure.
- The master internal instance model contains both structural parent object instances and component object instances that were generated from the entity arguments of the main predicate expression. E.g. for the councilmen and demonstrators schema, object instances include an object instance for each of *councilmen*, *demonstrators*, and *permit*.
- The master instance model also contains instantiations of the main meaning unit predicate-based behavior, based on a higher behavior class (the first found behavior class can be used here as it contains commonly-shared information for all behavior classes). E.g. for the councilmen and demonstrators schema, object instances will have had *state attribute* values set based on the definition of a "refusing something due to fear" behavior class.

**current predicate expression**: contains one or more unresolved pronouns:

- It is described by the *pronoun feature set* data structure rather than the spanning information data structure.
- The master internal instance model contains partial information based on common nouns, proper nouns, any resolvable pronouns (e.g. possessive pronouns), and verb information that can be determined prior to resolution of the unresolved pronoun.

The processing of the current predicate expression by the semantic engine is the main focus of the algorithms of the present method; however several other engine preparation tasks are also described.

### 5.3.2. Entity Resolution (Class Selection)

The task that involves selection of a relevant class from the ontology for a noun head word or noun phrase is referred to herein as "entity resolution"; this may to some extent overlap with the common usage of "word sense disambiguation". The syntactic and SNF (semantic) information about a word or phrase may provide sufficient constraints to allow unambiguous resolution (e.g. where "man" contained in a noun phrase constrains the ontology lookup process to object frame classes; behavior classes in the ontology are not searched). Other aspects of entity resolution that are important but not addressed here include:

- examination of the immediate in-sentence context to determine attribute types and behaviors that apply to one candidate class but not another or others
- probabilistic approaches
- commonsense inferences that can be triggered

The entity resolution task resolves a word or phrase by associating it with a class in the ontology. This involves a set of assertions – i.e. the features of the class - about the entity that is described by the word or phrase. Refer to Hofford (2014 (b)) "The ROSS User's Guide and Reference Manual" for detail about the object frame class.

### 5.3.3. Generation of Internal Instance Model

The semantic engine executes two main tasks that are part of master internal instance model generation. They are:

- Object instance instantiation: this occurs after entity resolution/class selection. Structural parent object instances and component object instances are inserted into the master instance model.
- Behavior class selection and application: this usually occurs after object instance instantiation and involves the application of a behavior class in order to generate additional information that can be determined. A typical case of behavior class application, as it would apply to the meaning unit “Bob hit the other man” would add additional attribute information to objects in the instance model (Bob, other man), and would also generate a new structural parent object instance at a separate point along a time-line: this new structural parent instance will hold cloned copies of the object instances (Bob, other man) and these object instances will have state attribute values that represent the results of the “hit” action.

### 5.3.4. Other Engine Tasks

The following tasks are also performed by the engine. Details about pronoun resolution and the embedded inference process will be provided in the following sections.

- Pronoun resolution: this takes place within entity resolution but also performs object instance instantiation (object instances based on pronouns can be instantiated as soon as the pronoun antecedent is resolved, therefore this task is incorporated into the final stages of the pronoun resolution task).
- Embedded inference/commonsense reasoning: this is performed when additional instance model information is needed. Types of inference include the following (triggered inference is not covered in this document)
    - Inferences that can be triggered by information that is gained as the input natural language text is processed. E.g. a sentence in a story provides information that gets used to create an instance model, from which an inference can be drawn: “As the mule slowly descended the rocky trail, suddenly it lost its footing and fell into the vast open space below”. (Triggered inference: the mule got injured or killed).
    - Inferences that are used within the pronoun resolution routine in order to handle the pronoun resolution task where other simpler instance-model-based attempts have failed. This is described in a subsequent section and has been applied in order to solve Winograd schema #1 (councilmen and demonstrators) for the “advocate violence” variant.
- Generation of external instance model: this is an optional step that involves generation of an external XML-based instance model.
- Question answering: the Comprehendor NLU system stores instance model information, which is used for follow-up question answering.

# 6. Semantic Engine Driver: Data Structures and Control Flow

## 6.1. Overview

A semantic engine sub-system for the present method will have a high-level function that is referred to as the *engine driver*. The EngineDriver() function processes an input Document, which is a list of communication units. The engine driver branches to an appropriate subroutine, depending on whether the communication unit is part of a sentence or is of another communication unit type, e.g. a web URL or a standalone email address. When it has finished processing all communication units, the engine driver invokes GenerateOutputInstanceModels() in order to generate the external (XML) version of the internal instance model and any other selected output forms (e.g. a bullet-point summary of a story).

Data structures and code from the Comprehendor NLU system are used to illustrate engine driver concepts. Comprehendor is a C++ implementation; the following sections use C++ or pseudo-code. *(Note: data structures shown here and the functions that follow are not intended as full listings of the actual code in the Comprehendor system).*

## 6.2. Data Structure for Master Token List

The engine requires an input master list of lexical tokens: the main data structure that is part of the implementation of this token list, referred to in the following sections, is "TokenListNode" (a list node class).

## 6.3. Data Structures and Data Types: Input to the Engine

This section describes the following data structures and their supporting definitions: *communication unit*, *sentence, predicate expression*, *predicate specifier*, *entity argument specifier*, and *modification specifier*.

A **communication unit** data structure has the following structure:

```
class CommunicationUnit
{
public:
   enum CommunicationUnitType communicationUnitType;

   // pointers to start and end tokens in the master input token list:

   TokenListNode *pFirstTokenListNode;
   TokenListNode *pLastTokenListNode;  // (e.g. for a declarative sentence, points to period token)

   Sentence *pSentence;
}
```

The CommunicationUnitType enumerated type is as follows (other types may be defined as needed)

```
enum CommunicationUnitType
{
   CommunicationUnitTypeSentence = 0,
   CommunicationUnitTypeURL,
   CommunicationUnitTypeEmailAddress,
   CommunicationUnitTypeSingleWordOnLine,
   CommunicationUnitTypeTwoWordPhraseOnLine,
   CommunicationUnitTypeAuthorInfo,
   CommunicationUnitTypeNONE   // max value for this enum
};
```

A **sentence** has the following structure; note that semicolon expressions are handled as top-level expressions within a sentence as they are similar to full sentences. (This structure also shows flags relating to paragraphs and quotations that have uses that are not described in detail here).

```
class Sentence
{
public:
   char szContentString [MAXLEN_CONTENTSTRING];  // stores the original sentence

   DiscourseContext discourseContextMajor;

   bool fSentenceStartIsParagraphBegin;
   bool fSentenceEndIsParagraphEnd;
   bool fSentenceIsPrependedWithQuotationBegin;
   bool fSentenceIsAppendedWithQuotationEnd;

   // if it contains any semicolon expressions:
   SemicolonExpressionNode *pSemicolonExpressionHeadNode;

   // if it does not contain any semicolon expressions:
   PredicateExpressionList predicateExpressionList;
}
```

The DiscourseContext enumerated type is defined as follows. A DiscourseContext value roughly corresponds to the grammatical concepts of tense and aspect, with further divisions that are partly based on mood. ("declarative" and "interrogative", although both are indicative, are separated here due to the needs of the semantic engine).

```
enum DiscourseContext
{
   DeclarativePastSimple = 0,
   DeclarativePastPerfect,
   DeclarativePastProgressive,
   DeclarativePastPerfectProgressive,

   DeclarativePresentSimple,
   DeclarativePresentPerfect,
   DeclarativePresentProgressive,
   DeclarativePresentPerfectProgressive,

   DeclarativeFutureSimple,
   DeclarativeFuturePerfect,
   DeclarativeFutureProgressive,
   DeclarativeFuturePerfectProgressive,
```

```
    InterrogativePastSimple,
    InterrogativePastPerfect,
    InterrogativePastProgressive,
    InterrogativePastPerfectProgressive,

    Imperative,

    Hypothetical,  // (related to the subjunctive mood)

    DiscourseContextNONE
};
```

The predicate expression list is not shown; a **predicate expression** has the following structure. The PredicateExpressionPointerList is not shown: this is a list of pointers to predicate expressions (including the current PE) that represents the original syntactic order of meaning units. For instance, given the sentence “The man could not lift his son because he was so weak.”, the pointer list points to two PEs that represent the two meaning units that are shown here:

- (head of list) “The man could not lift his son” // pointer to *this* data object
- (next/tail) “because he was so weak”  // pointer to a PE that is nested within the ModificationSpecifierList

The TokenListNode data structure is not shown: this is for a master token list that is generated by a lexical analyzer; token list nodes also store disambiguation information as it is determined by the engine. The pFirstTokenListNode pointer points into the master token list at the location that corresponds to the start of the predicate expression (this is usually the start token of a sentence).

```
class PredicateExpression
{
public:
    GrammaticalMood grammaticalMood;
    char szIntroductoryWord [MAXLEN_SINGLEWORDSTRING];   // e.g. that
    PredicateSpecifierList  predicateSpecifierList;
    EntityArgumentSpecifierList  entityArgumentSpecifierList;
    AttributiveArgumentSpecifierList attributiveArgumentSpecifierList;
    ModificationSpecifierList  modificationSpecifierList;
    PredicateExpressionPointerList  predicateExpressionPointerList;  // list order represents original syntactic
order of MUs
    TokenListNode *pFirstTokenListNode; // start token only; end token is marked and does not need to be
stored
}
```

The GrammaticalMood enumerated type is defined as follows.

```
enum GrammaticalMood
{
    GrammaticalMoodIndicative = 0,
    GrammaticalMoodInterrogative,
    GrammaticalMoodImperative,
    GrammaticalMoodNONE
};
```

The predicate specifier list is not shown; the **predicate specifier** is defined as follows:

```
class PredicateSpecifier
{
   int ordinal;
   char szMainVerbWord [MAXLEN_SINGLEWORDSTRING];    // e.g. walked, walking
   // char szParticleWord [MAXLEN_SINGLEWORDSTRING];    // e.g. up, out, over, in
   PredicateSpecifierRole semanticRole;
   DiscourseContext discourseContextActual;
   char szTrailingConnectiveWord[MAXLEN_SINGLEWORDSTRING];    // e.g. “and”
};
```

The PredicateSpecifierRole enum is as follows.

```
enum PredicateSpecifierRole
{
   PredicateToBeAttributive,                 // “The sky is gray.”
   PredicateToBeIsA,                         // “A car is a vehicle.”
   PredicateCapability,                      // “can”
   PredicateHasAVerb,                        // “A vehicle has wheels.”
   PredicateToBeTakingEntityArgument  // “The car is in the garage.”
   PredicateVerbTakingEntityArgument  // “The man walked.”
};
```

The entity argument specifier list is not shown; the **entity argument specifier** is shown here. Note that the entity argument specifier may contain a predicate expression, allowing for recursivity/nesting of predicate expressions. (Refer to the section above on Semantic Normal Form for definition of the EntityArgumentSemanticRole enumerated type).

```
class EntityArgumentSpecifier
{
   EntityDesignatorList  entityDesignatorList;  // (empty if unused)
   PredicateExpression *pPredicateExpression;  // (NULL if unused)
   EntityArgumentSemanticRole  semanticRole;  // one of: Actor, Actee, Extra
   ExtraSubRole  extraSubRole;
   SyntacticRole  syntacticRole;   // e.g. subject, direct object, indirect object
   int ordinalPredicate; // refers to a predicate specifier
};
```

The entity designator list is not shown; the **entity designator** is defined as follows:

```
class EntityDesignator
{
   NounPhrase *pNounPhrase;
   PrepositionalPhraseComplement *pPrepositionalPhraseComplement;
   char szTrailingConnectiveWord[MAXLEN_SINGLEWORDSTRING];    // e.g. “and”
};
```

The attributive argument specifier data structures are not shown here. The modification specifier list is not shown; the **modification specifier** is as follows. Note that the modification specifier may contain a predicate expression, allowing for recursivity/nesting of predicate expressions.

```
class ModificationSpecifier
{
   AdverbialPhrase  *pAdverbialPhrase;              // e.g. "quickly"
   AdverbialExpression  *pAdverbialExpression;  // e.g. "I awoke while it was still dark."
   PredicateExpression *pPredicateExpression;   // e.g. "The snows came early that year,
                                                 //          driving the bears into an early hibernation."
   SyntacticPosition  syntacticPosition;   // e.g. Leading, PreVerb, PostVerb, Final
   int ordinalPredicate; // refers to a predicate specifier
};
```

### 6.4. Data Structures and Data Types: Internal/Operational

The *ObjectInstance* structure is shown here: this is the main data structure that is used for information within an internal instance model. *(Note: this shows one of two alternatives (fixed-length array) for storing attributes and relationships – the second approach uses a list).*

```
class ObjectInstance
{
private:
   ObjectFrameClass *m_pReferenceObjectFrameClass;   // (ptr to class from which it was instantiated)

public:
   char szContentString [MAXLEN_CONTENTSTRING_STAR];  // e.g. stores "councilmen"

   char szUniqueIdentifier [MAXLEN_UNIQUEID_STRING];  // unique id

   bool fMultiple;  // specifies this as a collection (set) of object instances

   //-------------------------------------------------------
   // Features of the Object Instance:
   //
   //    - upon instantiation, each of the following is derived using any
   //       available features from the object frame class.
   //    - during subsequent instance model generation, new features may be added.
   //
   RelationshipToParent relationshipToParent;

   // (from ObjectFrameClass::Structure structure)
   InstanceStructure structure;

   // (from ObjectFrameClass::Attributes attributes)
   AttributeBaseExpression *rgpAttributeExpressions [MAX_OBJECTFRAMEINSTANCE_ATTRIBUTES];

   // (from ObjectFrameClass::Relationships relationships)
   RelationshipExpression *rgpRelationshipExpressions [MAX_OBJECTFRAMEINSTANCE_RELATIONSHIPS];

   //-------------------------------------------------------
   // List of associated behaviors:
   //
   BehaviorClassList behaviorClassList;

   // Methods not shown
};
```

The *InstanceStructure* member contains embedded objects: this is of particular importance for instance models, insofar as a structural parent object instance is only a "holder". (Structural parent object instances exist at the top level in an instance model as members of Contexts, described later). For instance, a structural parent instance based on the EverydayObjectStructuralParentClass may contain an object instance for a HouseClass and a DrivewayClass. An analogy that may help illustrate these concepts is the diorama: a structural parent object instance is like a diorama that is frozen at one instant of time; the object instances that it contains (e.g. a house, a car) are like objects in a diorama. (The representation of time is accomplished by the use of the Context).

The *ObjectInstanceSemanticWrapper* structure is used by the spanning information data structure:

```
class ObjectInstanceSemanticWrapper
{
   // (reference-only: do not deallocate this pointer)
   ObjectInstance *pObjectInstance;

   //--------------------------------------------------------
   //  SNF Information:
   //
   EntityArgumentSemanticRole  semanticRole;  // one of: Actor, Actee, Extra
   ExtraSubRole  extraSubRole;
   SyntacticRole  syntacticRole;   // e.g. subject, direct object, indirect object
   int ordinalOfPredicate;
};
```

The *BehaviorClassesPerMainVerbWrapper* stores a main verb word and the associated behavior classes. E.g. given the sentence, "The man could not lift his son or carry his daughter because he was too weak.", this stores information to relate a list of behavior classes for each predicate separately:

```
class BehaviorClassesPerMainVerbWrapper
{
   char szMainVerbWord [MAXLEN_SINGLEWORDSTRING];    // e.g. "lift", "carry"
   BehaviorClassList  behaviorClassList;  // e.g. NotPersonLiftsPerson01, NotPersonLiftsPerson02
};
```

There are several *ActivePointer* structures that are maintained per predicate expression and that are used internally by the engine: they are not shown here. The *SpanningInformation* structure contains information that normally corresponds to the previous predicate expression; it stores pointers to object instances in the master internal instance model.

```
struct SpanningInformation
{
   DiscourseContext discourseContextSaved;

   // (do not call delete for these pointers)

   Context *pContextMRU;  // (most-recently-used Context in the master instance model)

   ObjectFrameClass *pObjectFrameClassStructuralParent;
```

```
   ObjectInstance *pObjectInstanceStructuralParent;

   // The main list of ObjectInstanceSemanticWrappers:
   //
   ObjectInstanceSemanticWrapperList  objectInstanceSemanticWrapperList;

   BehaviorClassPerMainVerbWrapperList  behaviorClassPerMainVerbWrapperList;

   // (Methods not shown)

};  // struct SpanningInformation
```

The *SpanningInfoStack* allows for the storage of multiple spanning infos. SpanningInformation pointers (referred to as "spanning info's") are pushed onto the stack in an order that is dependent on the control strategy for processing of PEs (by default this order reflects the order of original syntactic meaning units). The engine uses a stack trim operation (not shown) in order to limit the size of the stack: this is based on the heuristic assumption that there is a limit to the number of prior sentences/clauses that may intervene between an antecedent referent and an anaphoral pronoun. E.g. the stack trim may be invoked so that the size of the stack stays in the range of 10 to 15 meaning units.

```
//---------------------------------------------------------------------------
//
//  class SpanningInfoStack
//
//---------------------------------------------------------------------------
//
struct SpanningInfoStackNode
{
   // Data:
   SpanningInformation *pSpanningInformation;

   SpanningInfoStackNode *up;
   SpanningInfoStackNode *down;

   // (Methods not shown)
};

class SpanningInfoStack
{
private:
   SpanningInfoStackNode *curr;

public:
   SpanningInfoStackNode *top;

   // Methods:
   SpanningInfoStack ();
   void Push (SpanningInformation *pSpanningInformation);
   bool Pop (SpanningInformation **ppSpanningInformation);
   bool Current (SpanningInformation **ppSpanningInformation);
   void ResetCurrentToTop ();
   void DiscardAll();
```

```
};  // struct SpanningInfoStack
```

The *PronounFeatureSet* data structure stores all information that can be gathered about the pronoun and its context within the clause in which it appears. Several supporting enumerated types are shown first:

```
//---------------------------------------------------------------------------
//
//  enum PredicateExpressionTemporalOrderIndicator
//
//---------------------------------------------------------------------------
//
enum PredicateExpressionTemporalOrderIndicator
{
   PredicateExpressionTemporalOrderIndicatorFollowing,
   PredicateExpressionTemporalOrderIndicatorPreceding,  // e.g. for "after"
   PredicateExpressionTemporalOrderIndicatorUndetermined,
   //
   PredicateExpressionTemporalOrderIndicatorNONE
};

//---------------------------------------------------------------------------
//
//  enum PredicateExpressionHypotheticalUsage
//
//    Note: a predicate expression may have a hypothetical usage ("because ...")
//          and at the same time convey declarative information.
//
//---------------------------------------------------------------------------
//
enum PredicateExpressionHypotheticalUsage
{
   PredicateExpressionHypotheticalUsageExplanationOfCause,  // "because"
   PredicateExpressionHypotheticalUsageExplantionOfEffect,      // "causing ..."
   PredicateExpressionHypotheticalUsageExplantionOfObjective,   // "in order to ..."
   //
   PredicateExpressionHypotheticalUsageNONE
};

//---------------------------------------------------------------------------
//
//  enum PronounGender
//
//---------------------------------------------------------------------------
//
enum PronounGender
{
   PronounGenderMale = 0,
   PronounGenderFemale,
   PronounGenderNonspecific,
   //
   PronounGenderNONE
};

//---------------------------------------------------------------------------
//
```

```
//  enum PronounCardinality
//
//---------------------------------------------------------------------------
//
enum PronounCardinality
{
   PronounCardinalitySingular = 0,
   PronounCardinalityPlural,
   PronounCardinalityNonspecific,
   //
   PronounCardinalityNONE
};

//---------------------------------------------------------------------------
//
//  enum PronounActiveOrPassive  // (e.g. "they" (active) versus "them" (passive))
//
//---------------------------------------------------------------------------
//
enum PronounActiveOrPassive
{
   PronounActiveOrPassiveActive = 0,
   PronounActiveOrPassivePassive,
   PronounActiveOrPassiveNonspecific,
   //
   PronounActiveOrPassiveNONE
};

//---------------------------------------------------------------------------
//
//  enum SyntacticRole
//
//---------------------------------------------------------------------------
//
enum SyntacticRole
{
   SyntacticRoleSubject = 0,
   SyntacticRoleDirectObject,
   SyntacticRoleIndirectObject,
   //
   SyntacticRoleNONE
};

//---------------------------------------------------------------------------
//
//  enum SemanticRole
//
//---------------------------------------------------------------------------
//
enum SemanticRole
{
   SemanticRoleActor = 0,
   SemanticRoleActee,
   SemanticRoleExtra,
   //
   SemanticRoleNONE
};
```

```
//-----------------------------------------------------------------------------
//
//  struct PronounFeatureSet
//
//-----------------------------------------------------------------------------
//
struct PronounFeatureSet
{
   char szPronounWord [MAXLEN_SINGLEWORDSTRING];

   PronounGender pronounGender;
   PronounCardinality pronounCardinality;
   PronounActiveOrPassive pronounActiveOrPassive;

   // PredicateExpression Fields:
   PredicateExpressionTemporalOrderIndicator predicateExpressionTemporalOrderIndicator;
   PredicateExpressionHypotheticalUsage predicateExpressionHypotheticalUsage;

   DiscourseContext discourseContext;

   SyntacticRole syntacticRole;
   SemanticRole semanticRole;

   //=======================================
   // Other Object Instances:  (non-pronouns or resolved pronouns)
   //
   ObjectInstanceSemanticWrapperList  objectInstanceSemanticWrapperList;
   //=======================================

   bool fNegationOfSearchKeyWord;

   PredicateSpecifierRole predicateSpecifierRole;

   char *pszSearchKeyWordAdjective;   // non-null for AttributiveArgumentRole = Attribute

   char *pszSearchKeyWordVerb;  // e.g. "refused"

   // (Methods not shown)

};  // struct PronounFeatureSet
```

## 6.5. Data Structures and Data Types: Instance Model

The instance model is implemented as a ContextList. The following listing contains the Context data structure and the ContextList structure. The map that contains all top-level structural parent instances is in bold:

```
struct Context
{
   char szUniqueIdentifier [MAX_SIZE_UNIQUEID_STRING];
   DiscourseContext discourseContext;
   char szLeadingObjectInstanceClassName [MAXLEN_CONTENTSTRING_STAR];
   char szTemporalAttributeValueLastUsed [ATTRIBUTE_VALUE_MAX_SIZE];
```

```
   //-----------------------------------------------------------------------------------------------------
   // Map that contains all structural parent instances, indexed by time attributes:
   //
   MapObjectInstances *pMapObjectInstances;

   // Methods not shown:
};

struct ContextListNode
{
   Context *pContext;

   struct ContextListNode *prev;
   struct ContextListNode *next;

   // Methods not shown:
};

class ContextList
{
private:
   ContextListNode *m_head;
   ContextListNode *m_tail;

public:
   // Public methods not shown:
};
```

The MapObjectInstances structure stores structural parent object instances, each of which is indexed by a temporal attribute value.

```
// MapObjectInstances:
//
//  - the wrapper class is not shown; the map of object instances contains ObjectInstance pointers:
//
typedef map <string, ObjectInstance*> MapTypeObjectInstances;
typedef pair<MapTypeObjectInstances::iterator,bool> retvalMapTypeObjectInstances;
```

## 6.5. Engine Driver Algorithm

The following is the control flow for the engine driver that iteratively processes all communication units and does several other tasks. During processing, the engine driver builds the master internal instance model, generates spanning information within the spanning info stack, and also adds disambiguation information, as it is determined, to the master token list that is associated with the syntactic and semantic normal form information. It also generates an external instance model, and any other external information artifacts (e.g. bulleted summary) that have been specified.

```
EngineDriver  (  input:  list of communication units (containing embedded PEs using SNF)
                 output: disambiguation information (stored in master token list),
                 output: instance model and other selected models )

   // other tasks here not shown
```

```
ProcessAllCommunicationUnits ()

GenerateOutputInstanceModels ()

Return
```

ProcessAllCommunicationUnits() is shown next. For each communication unit that is a sentence, this routine extracts a pointer to the first predicate expression in the list and passes it to the function **ProcessPredicateExpression()** (shown inline). The iterative processing performs the same task for all subsequent predicate expressions within the same list within the sentence.

```
ProcessAllCommunicationUnits ()

  pCommUnitList->First()

  while (!pCommUnitList->IsDone())
  {
    if (pCommUnit->communicationUnitType != CommunicationUnitTypeSentence)
      // Handle other comm unit types and continue …

    Sentence *pSentence = pCommUnit->pSentence;

    PredicateExpression *pPredicateExpression = pSentence->GetFirstPredicateExpression();

    while (pSentence->IsDonePredicateExpressionList())   // (loop to get all predicate expressions)
    {
      //  process one predicate expression:

      if (pPredicateExpression != NULL)

        ProcessPredicateExpression ()
        {
          switch (pPredicateExpression->grammaticalMood)
          {
          case GrammaticalMoodIndicative:

             ProcessPredicateUnitIndicative ()

          case GrammaticalMoodInterrogative:

            ProcessPredicateUnitInterrogative ()

          case GrammaticalMoodImperative:

            ProcessPredicateUnitImperative ()
          }
       }

      pPredicateExpression = pSentence->GetNextPredicateExpression ();

    }   // while  (loop to get all predicate expressions)

    pCommUnitList->Next()

  }  // while (loop to get all communication units)
```

```
Return // from ProcessAllCommunicationUnits ()
```

**ProcessPredicateUnitIndicative** processes a predicate expression as follows. The details of the control strategy for populating and maintaining the spanning information stack are not provided here; by default, the spanning information stack items get pushed onto the stack in the order in which they appear within the input natural language text. The PredicateExpression::PredicateExpressionPointerList is used by default to support population of the spanning info stack using a sequence that corresponds to the syntactic order of the original input meaning units. For instance, a leading adverbial clause such as "Before the first light dawned", will generate a spanning info that gets pushed onto the stack before the subsequent main clause with the same sentence. (E.g. full sentence: "Before the first light dawned, Joe ran several miles.").

Because the main control strategy within this function is iterative, and indirect recursive calls are used to process nested PEs, there is a limit to the levels of nesting in the input that are processed recursively. Examples:

- "The trophy *that Tim won* doesn't fit in the suitcase." – one level of nesting, processed by a recursive call.
- "The trophy *that Tim won for the contest* ***that he entered last year*** doesn't fit in the suitcase." – two levels of potential recursion, however the second bound relative clause is not handled by a nested recursive call; rather "that he entered last year" is processed in the iteration sequence since it is pointed to by a node in the PredicateExpressionPointerList.

(Note: the functionality of ProcessPredicateUnitInterrogative() and ProcessPredicateUnitImperative() are similar and are not shown here).

```
ProcessPredicateUnitIndicative (IN: PredicateExpression *pPredicateExpressionMain,
                                OUT: SpanningInfoStack *pSpanningInfoStack,
                                OUT: InstanceModel *pInstanceModel)

   SetTimelineHintInformation ()   // examine adverbial information that indicates time and temporal order

   PredicateExpressionPointer *pPredicateExpressionPointer = pPredicateExpressionMain->GetFirstPEPointer();

   while (!pPredicateExpressionMain->IsDonePredicateExpressionPointerList())
   {
      If (pPredicateExpressionPointer points to self/this)
      {
         // (not shown: process arguments for predicate type PredicateToBeAttributive; new attribute
         //   information gets added to the appropriate instance model object instance)

         // (not shown: process arguments for each of predicate types: PredicateToBeIsA, PredicateHasAVerb)

         // Process non-nested entity argument, e.g. "Joe", "miles" that do not contain pronouns:

         ProcessNonPronounEntityArguments (pPredicateExpressionMain-> entityArgumentSpecifierList,
                                           pSpanningInfoStack,
                                           pInstanceModel );

         ProcessPronounEntityArguments (pPredicateExpressionMain-> entityArgumentSpecifierList,
                                        pSpanningInfoStack,
```

```
                                             pInstanceModel );

      if (failed to resolve the pronoun)
      {
         // Lookahead:  (possibly a cataphoric pronoun)
         //
         //    E.g. "Because it was too big, the trophy did not fit in the suitcase."

         pPredicateExpressionPointer = pPredicateExpressionMain->GetNextPEPointer();

          // process non-nested entity arguments in the PE, populate a temporary spanning info

          // retry: ProcessPronounEntityArguments()  // use temp spanning info
      }

      // Extract predicate adverbials: get adverbs and adverb phrases that modify a verb
      // in the predicate specifer, (e.g. "not", e.g. "quickly")

      ExtractPredicateAdverbials ();

      if (predicateSpecifierRole == PredicateVerbTakingEntityArgument)
      {
           ProcessPredicateSpecifierList ();
      }
   }
   Else
   {
      // The following will do indirect recursion to process the PE pointed to by pPredicateExpressionPointer:

      If (nested PE is within a modification specifier)
       {
          ProcessModificationSpecifier ();     // e.g. leading adverbial phrase, e.g. final adverbial phrase
      }
      Else if (nested PE is within an entity argument)
       {
          ProcessNestedEntityArgument ();  // e.g. gerundive phrase, e.g. "walking to the corner"
                                           // e.g. nested bound relative clause, e.g. "the person who fell ill"
      }
   }

   pPredicateExpressionPointer = pPredicateExpressionMain->GetNextPEPointer();

}  // while (loop to get top-level and nested PEs from pPredicateExpressionMain)

Return
```

**ProcessNonPronounEntityArguments ()** iteratively process the entity arguments that do not contain pronouns:

```
ProcessNonPronounEntityArguments  (IN: entityArgumentSpecifierList,
                                   OUT: pSpanningInfoStack,
                                   OUT: pInstanceModel)

while (entityArgumentSpecifierList is not empty)
{
   If (entity argument specifier does not contain a pronoun)
```

```
        ProcessEntityArgument (pEntityArgumentSpecifier, pSpanningInfoStack, pInstanceModel)

    }  // while ()

Return
```

**ProcessEntityArgument** () **-** this function processes information that was originally part of the syntactic subject, direct object, indirect object and any other entities within prepositional phrase complements within the meaning unit that corresponds to the predicate expression. The entity resolution (class selection) tasks for the actors, actees, and extras of the predicate unit are performed within ProcessEntityArgument (). For all noun phrases except those that contain unresolved pronouns, internal instance model instantiation takes place. *(MainDriverForInstanceModelGeneration() represents a complex sub-system, the functionality of which is outside the scope of this document).* The control flow of ProcessEntityArgument () is as follows (showing the lower level functions inline).

```
ProcessEntityArgument ( IN: EntityArgumentSpecifier *pEntityArgumentSpecifier,
                        OUT: pSpanningInfoStack,
                        OUT: pInstanceModel)

  // EntityArgumentSpecifier:
    // EntityDesignatorList  entityDesignatorList;
      // EntityDesignator
        // NounPhrase *pNounPhrase;
        // PrepositionalPhrase *pPrepositionalPhrase;
        // char szTrailingConnectiveWord; // e.g. "and"
    // EntityArgumentSemanticRole  semanticRole;  // one of: Actor, Actee, Extra
    // ExtraSubRole  extraSubRole;
    // SyntacticRole  syntacticRole;   // e.g. subject, direct object, indirect object
    // int ordinalPredicate; // refers to a predicate specifier


  Loop:   // process each item in entityDesignatorList:

    switch (type)  // noun phrase or prepositional phrase
    {
    case NounPhrase:
      ProcessNounPhrase ()
      {
            EntityResolutionRoutine (pEntityDesignator->pNounPhrase,
                                     pSpanningInfoStack)
      }
      break;

    case PrepositionalPhrase:
      ProcessPrepositionalPhrase ()
      {
            // (note: if role is Extra, SubRole is available here)

            // Extract noun phrase (not shown)

            EntityResolutionRoutine (pEntityDesignator->pPrepositionalPhrase->pNounPhrase,
                                     pSpanningInfoStack)
      }
```

```
      break;

    } // switch

    MainDriverForInstanceModelGeneration ()    // Instance Model Generation (not shown)

  End Loop

 Return
```

**ProcessPronounEntityArguments ()** is similar to ProcessNonPronounEntityArguments(); it calls **ProcessPronounEntityArgument ()**, which is similar to ProcessEntityArgument(), and only processes entity arguments that contain pronouns. (Detail not shown).

**ProcessModificationSpecifier ()** is shown next; the main processing task is to handle the nested predicate expression, for which ProcessPredicateExpression() is invoked.

```
ProcessModificationSpecifier (IN: pPredicateExpression,
                              OUT: pSpanningInfoStack)

  // other tasks here not shown

  ProcessPredicateExpression (pPredicateExpression, pSpanningInfoStack)

Return
```

**ProcessPredicateSpecifierList()** is outlined here. Like ProcessEntityArgument(), this function also generates instance model information, but it is based on main verbs that represent processes (it is only invoked for predicateSpecifierRole == PredicateVerbTakingEntityArgument). (Note: C++ arguments are shown for some calls to lower-level routines).

```
ProcessPredicateSpecifierList ()

  pPredicateSpecifier = PredicateSpecifierList->First();

  while (!PredicateSpecifierList ->IsDone())
  {
    // Search to get list of all relevant behavior classes:

    iResult = SearchObjectFrameClassBehaviorClasses (pPredicatePhrase->szMainVerbWord,
                              NULL, // word2
                              NULL, // word3
                              RuleDirectionUnspecified,
                              fNegation,
                              pActivePointersForClasses->clustSem.pObjectFrameClassActorList,
                              pActivePointersForClasses->clustSem.pObjectFrameClassActeeList,
                              pActivePointersForClasses->clustSem.pObjectFrameClassExtraList,
                              &pBehaviorClassList);
    // check iResult …

    // Apply/process the behavior class (first behavior class or higher behavior class)

    //  (this performs possibly extensive Instance Model Generation  (not shown))
```

```
        iResult = ProcessBehaviorClassForSubjectActive (pInfopedia,
                                        pActivePointersForClasses,
                                        pActivePointersForInstances,  // (contains linkage into master instance model)
                                        pBehaviorClassList->GetFirstBehaviorClass());
        // check iResult …

        // Completion tasks for Instance Model Generation (not shown)

        // Set SpanningInfo fields (not shown)

        pPredicateSpecifier = PredicateSpecifierList ->Next();
    }

Return
```

## 7. General Pronoun Resolution Method

This section explains the general pronoun resolution method. This method has been used to process the following Winograd Schema Challenge schemas:

- trophy and suitcase schema (this schema is processed using the general pronoun resolution method as it is supplemented by the modeling of the communicative agent (see *Appendix 1: Solution for "Trophy and Suitcase" Schema Using a Model of the Communicating Agent*).
- man cannot lift his son
- Joe paid the detective
- city councilmen refusing a permit because they feared violence.

A subsequent section describes the *embedded commonsense reasoning method* that is invoked during execution of the general pronoun resolution method when a situation is encountered that cannot be solved by the general method. The embedded commonsense reasoning method is applied when resolving this schema:

- city councilmen refusing a permit because they advocated violence.

### 7.1. Entity Resolution (Class Selection)

This function is called EntityResolutionRoutine (). For situations where a pronoun exists in any meaning unit/predicate expression other than the first within the input text, this function will get invoked twice:

- During processing of the main predicate expression
- During processing of the current predicate expression

```
EntityResolutionRoutine (IN: EntityArgumentSpecifier *pEntityArgumentSpecifier,
                        IN/OUT: ActivePointers *pActivePointers,
```

```
                                    OUT: SpanningInfoStack *pSpanningInfoStack)
{

   // Loop: iterate through the list of noun phrase head words: e.g. "Fred and Mary walked their dog."

   NounHeadWordListNode *pNounHeadWordCurrNode = pNounPhrase->pMainWordHeadNode;

   while (pNounHeadWordCurrNode != NULL)
   {
      switch (pNounHeadWordCurrNode->nounWordPhraseType)
      {
         case NounWordPhraseTypePronoun:
            ProcessPronoun (pSpanningInfoStack, pActivePointers, pPronounFeatureSet);
         break;

         case NounWordPhraseTypeExistentialThere:
            ProcessExistentialThere();
         break;

         case NounWordPhraseTypeCommonNoun:
            ProcessCommonNounPhrase ();

         case NounWordPhraseTypeProperNoun:
            ProcessProperNounPhrase ();

      }; // switch()

      pNounHeadWordCurrNode = pNounHeadWordCurrNode->next;

   }  // while ()

   // Determine a structural parent class that can be used for the entity or entities:

   iResult = GetBaseStructuralParentClass ();

Return from EntityResolutionRoutine()
```

The functions to process common nouns, proper nouns, and the existential “there” are not described here. ProcessPronoun() is described next.

### 7.2. ProcessPronoun() -> PronounResolutionGeneralMethod()

The ProcessPronoun() routine attempts to determine both the referent and the antecedent for a pronoun. Determination of the referent involves identifying both the object frame class and the specific instance model object instance. The spanning information stack is used throughout the functions that are invoked by ProcessPronoun(). An ActivePointers data object is also used (it is similar to the spanning information structure but it represents the *current* meaning unit).

ProcessPronoun() is not shown here: its primary function is to pass control to PronounResolutionGeneralMethod(), which is the main driver and worker routine for pronoun resolution. (The following functions show the C++ return types and several instances of error codes, such as E_SUCCESS and E_NOTFOUND).

```
int PronounResolutionGeneralMethod (SpanningInfoStack *pSpanningInfoStack,
                                    IN/OUT: ActivePointers *pActivePointers,
                                    PronounFeatureSet *pPronounFeatureSet,
{
   //---------------------------------------------------------------------------------
   // If pronoun is post-verb object (e.g. him/her/them/it) attempt to resolve it to a pre-verb entity
   //  (return if successful)
   //  (e.g. "The owners of the house sold it.")


   //---------------------------------------------------------------------------------
   //  Main driver for searching the instance model via the spanning information:
   //
   iResult = ExploratorySearchUsingSpanningInfoStack (pSpanningInfoStack,
                                                      pPronounFeatureSet,
                                                      &pMatchingObjectInstance,
                                                      // other parameters not shown);
   // check iResult …

   //------------------------------------------------------------------------
   // Check results from the exploratory search routine:
   //
   if (found a  referent)
   {
      *ppObjectFrameClassReferent = pMatchingObjectInstance->GetReferenceObjectFrameClass();

      // Save for disambiguation:
      pszOriginalObjInstWord = pMatchingObjectInstance->szContentString;  // "trophy", "Joe", "detective"
   }

   //  Modify the object instance within the (actual situation) instance model:

   SetAttributeWithinActualSituationEntity (pMatchingObjectInstance,
                                            szCausalFeatureAttributeType,  // e.g. "FunctionalAttribute1"
                                            szCausalFeatureValue);         // e.g. "TooSmall"

   //---------------------------------------------------------------------------------
   //  Process disambiguation tasks:
   //
   if (disambiguationTask == DisambiguationTaskPronouns)
   {
      iResult = SetPronounResolutionInformationInMasterTokenList (pszOriginalObjInstWord,
                                                                  pPronounFeatureSet,
                                                                  pFirstTokenNode);
      if (iResult != E_SUCCESS)
      {
         return iResult;
      }
   }  // if (disambiguationTask == DisambiguationTaskPronouns)

   return E_SUCCESS;

}  // PronounResolutionWorkerRoutine ()
```

## 7.3. SetPronounResolutionInformationInMasterTokenList ()

SetPronounResolutionInformationInMasterTokenList () inserts the results of the pronoun resolution into the master token list. *(The following feature not yet implemented: insert all words for a noun phrase, not just the noun head word (e.g. "trophy"))*.

```
int SetPronounResolutionInformationInMasterTokenList (char *pszOriginalObjInstWord,
                                                      PronounFeatureSet *pPronounFeatureSet,
                                                      TokenListNode *pFirstTokenNode)
{
   if (pszOriginalObjInstWord == NULL)
   {
      return E_NOTFOUND_REQUIREDITEM;
   }

   if (pPronounFeatureSet->szPronounWord[0] != '\0')
   {
      TokenListNode *pTokenNodeTEMP = pFirstTokenNode;

      // Search the token sub-list for the pronoun:

      while (pTokenNodeTEMP != NULL &&
           pTokenNodeTEMP->pMarkers->commUnitMarker != CommUnitMarkerEnd &&
           0 != strcmp (pTokenNodeTEMP->tokenvalue, pPronounFeatureSet->szPronounWord))
      {
         pTokenNodeTEMP = pTokenNodeTEMP->next;
      }

      if (pTokenNodeTEMP == NULL)
      {
         return E_NOTFOUND_REQUIREDITEM;
      }
      else
      {
         strcpy (pTokenNodeTEMP->tokenResolvedWord, pszOriginalObjInstWord);
      }
   }

   return E_SUCCESS;
}
```

## 7.4. ExploratorySearchUsingSpanningInfoStack()

ExploratorySearchUsingSpanningInfoStack () is a driver function that iterates to perform the pronoun resolution search logic against each spanning info in the spanning info stack.

```
int ExploratorySearchUsingSpanningInfoStack (
                                  // IN:
                                  SpanningInfoStack *pSpanningInfoStack,
                                  PronounFeatureSet *pPronounFeatureSet,
                                  // OUT:
                                  char *pszCausalFeatureAttributeType,
                                  char *pszCausalFeatureValue,
                                  bool *pfFoundPopObj,
                                  bool *pfFoundMatchingNestedBehavior,
                                  ObjectInstance **ppMatchingObjectInstance,
```

```
                                BehaviorClass **ppBehaviorClassNested)
{
   int iResult = E_NOTFOUND;

   // Main Loop: until reach bottom of StackOfSpanningInfos

   SpanningInformation *pSpanningInformation = NULL;

   pSpanningInfoStack->ResetCurrentToTop();

   while (pSpanningInfoStack->Current(&pSpanningInformation))
   {
      iResult = ExploratorySearchUsingOneSpanningInfo
                                pSpanningInformation,
                                pPronounFeatureSet,
                                pszCausalFeatureAttributeType,
                                pszCausalFeatureValue,
                                pfFoundPopObj,
                                pfFoundMatchingNestedBehavior,
                                ppMatchingObjectInstance,
                                ppBehaviorClassNested);
      if (iResult == E_SUCCESS)
      {
         break;  // Found
      }

   } // End Main Loop

   pSpanningInfoStack->ResetCurrentToTop();

   return iResult;
}
```

## 7.5. ExploratorySearchUsingOneSpanningInfo()

ExploratorySearchUsingOneSpanningInfo () iterates to test each candidate object instance in the instance model against the pronoun feature set. If multiple behavior classes are a match, the probabilities of the nested behaviors of each are compared in order to select the nested behavior and the object that have the highest probability.

```
ExploratorySearchUsingOneSpanningInfo (
                                // IN:
                                SpanningInformation *pSpanningInfo,
                                PronounFeatureSet *pPronounFeatureSet,
                                // OUT:
                                char *pszCausalFeatureAttributeType,
                                char *pszCausalFeatureValue,
                                bool *pfFoundPopObj,
                                bool *pfFoundMatchingNestedBehavior,
                                ObjectInstance **ppMatchingObjectInstance,
                                BehaviorClass **ppBehaviorClassNested)

   while (candidates exist)
   {
      TestOneCandidateObjectInstance (candidate)
   }
```

```
    // Compare Probabilities, allowing for the following to be set:  (compare logic not shown)

    *ppMatchingObjectInstance = pSpanningInformation->GetObjectInstance(idx);
    *ppBehaviorClassNested = pBehaviorClassNested[idx];
  }
```

## 7.6. TestOneCandidateObjectInstance()

TestOneCandidateObjectInstance () branches to an appropriate subroutine depending on the information that is available in the pronoun feature set:

- Adjective info: TryToMatchCausalFeatureAgainstSpanningInfoObjectInstance()
- Verb info: TryToMatchVerbCausalFeatureAgainstSpanningInfoObjectInstance()

(The logic of these two functions is not shown here).

If the input pronoun feature set contains a verb word, and if pronoun resolution fails after attempting to resolve it with TryToMatchVerbCausalFeatureAgainstSpanningInfoObjectInstance(), then GenerateAndTestForCausativeSituation() is invoked. This is the embedded sandbox-based generate and test method. Example of object instance candidates for W.S. schema #1 are "councilmen", "demonstrators", and “permit”. Here is the logic that is used when invoking the embedded reasoning routine:

```
TestOneCandidateObjectInstance ()

  // Test the candidate using TryToMatchVerbCausalFeatureAgainstSpanningInfoObjectInstance()

  if (iResult == E_NOTFOUND)
  {
    // GenerateAndTest: Try doing a forward inference for this candidate (e.g. "demonstrators")

    if (pPronounFeatureSet->meaningUnitHypotheticalUsage ==
                MeaningUnitHypotheticalUsageExplanationOfCause &&
      pPronounFeatureSet->pszSearchKeyWordVerb != NULL)
  {
    iResult = GenerateAndTestForCausativeSituation (pSpanningInformation,
                                                    pPronounFeatureSet,
                                                    pObjInstCandidate,
                                                    semanticRoleCandidate,
                                                    pBehavClassHeadNodeForCandidate,
                                                    pfFoundMatchingNestedBehavior,
                                                   ppBehaviorClassNested);
    if (iResult != E_SUCCESS && iResult != E_NOTFOUND)
    {
      return iResult;
    }
  }
```

GenerateAndTestForCausativeSituation() is the driver for the embedded reasoning tasks and is described in the next section.

## 8. Embedded Commonsense Reasoning Method

GenerateAndTestForCausativeSituation() is one of the main routines that implements the embedded commonsense reasoning method. This section will describe the embedded reasoning method using the example of WS schema #1/variant 2 ("advocated violence").

GenerateAndTestForCausativeSituation() is one of multiple possible routines – each of which performs inference for a similar purpose. GenerateAndTestForCausativeSituation() handles cases that involve finding a referent for which the antecedent exists in a clause (e.g. a "because" clause) that is an explanation of the cause for a situation. (Other such routines are not described in this document).

The main task of this function is that of testing the input object instance candidate (e.g. "the city councilmen", e.g. "the demonstrators", e.g. "a permit") to determine if it could have participated in a behavior that led to the fact that was stated in the main clause (e.g. "the city councilmen refused the demonstrators a permit"). Because it is not known at this stage whether or not the candidate is the referent (e.g. it was the demonstrators who advocated violence, not the councilmen), there is a need to create a temporary "sandbox" instance model that exists apart from the engine's master internal instance model. (Once the correct candidate has been identified, the master instance model will get updated with the newly-acquired information).

The sandbox instance model is a staging area that actually involves two separate instance model contexts: to avoid confusion these will be referred to as "east" side (earlier) and "west" side (later). The task of the routine is to build the two "sides" of the instance model and then determine whether or not they "meet up" using a final matching routine (similar to building an East railroad line and a West line and joining them in the middle). An overview of the process is as follows:

(West: represents earlier points along the time-line, starting with a candidate that might have "advocated violence") (the following is done iteratively for each main rule matching "advocated violence")

- given the object instance candidate, e.g. councilmen
- find a main rule (i.e. a rule for "advocate violence" – e.g. TalkerAdvocatesActionWithListenersWhoAnticipateSomething)
- create a sandbox context and insert an earliest structural parent instance into it
- apply the main rule to derive consequential information, save a pointer to a nested rule
- apply the nested rule (e.g. AnticipateHarmfulEvent) to generate new information into the West-side temporary context.

(East: represents later points along the time-line, working backward from the time point where "the city councilmen refused the demonstrators a permit") (the following is done iteratively)

- e.g. given "the city councilmen refused the demonstrators a permit" utilize master instance model information to create the East-side temporary context
- apply nested rule that is contained within the "refuse …" rule

- *(see subsequent section for logic that matches attribute state values from the West and East contexts)*

```
GenerateAndTestForCausativeSituation(
                        // IN:
                         PronounFeatureSet *pPronounFeatureSet,
                         ObjectInstance *pObjInstCandidate,
                         SpanningInformation *pSpanningInformation, // IN/OUT
                         BehaviorClassListNode *pBehavClassHeadNodeForCandidate, // list
                        // OUT:
                         bool *pfFoundMatchingNestedBehavior,
                         BehaviorClass **ppBehaviorClassNested)

// (example values for pBehavClassHeadNodeForCandidate; this parameter is not used here
//   but is passed into GenerateAndTest_ProcessOneForwardRule()
//      - RefusingSomethingDueToFearBehaviorClass
//      - RefusingSomethingDueToFearOnPartOfRequestorBehaviorClass)

//------------------------------------------------------------------
// Use pPronounFeatureSet->semanticRole to determine roles within current clause:
//
  // e.g. set pObjFrameClassActorTemp  (not shown)

//------------------------------------------------------------------
// SearchObjectFrameClassBehaviorClasses

  BehaviorClassListNode *pBehaviorClassListNodeHeadForwardRule = NULL;

  iResult = SearchObjectFrameClassBehaviorClasses
           (pPronounFeatureSet->pszSearchKeyWordVerb, // e.g. "advocated"
            NULL, // word2
            NULL, // word3
            RuleDirectionForward,
            false, // fNegationNestedBehavior,
            pObjFrameClassActorTemp,
            pObjFrameClassActeeTemp,
            pObjFrameClassExtraTemp,
            &pBehaviorClassListNodeHeadForwardRule);

  if (iResult != E_SUCCESS)
  {
    if (iResult == E_NOTFOUND_REQUIREDITEM)
    {
      return E_NOTFOUND;
    }
    return iResult;
  }

//------------------------------------------------------------------
// Loop:
//   - search each nested behavior class:  (e.g. " TalkerAdvocatesActionWithListenersWhoAnticipateSomething")
//
  BehaviorClassListNode *pBehavClassCurrNode = pBehaviorClassListNodeHeadForwardRule;

  while (pBehavClassCurrNode != NULL)
  {
    if (pBehavClassCurrNode->pBehaviorClass->pBehaviorClassExpression->fCausalRule)
```

```
        {
           iResult = GenerateAndTest_ProcessOneForwardRule(
                                   pSpanningInformation,
                                   pPronounFeatureSet,
                                   pObjInstCandidate,
                                   pBehavClassCurrNode->pBehaviorClass, // pBehaviorClassForwardRule
                                   pBehavClassHeadNodeForCandidate,
                                   pfFoundMatchingNestedBehavior,
                                   ppBehaviorClassNested);

           if (iResult == E_SUCCESS &&
              (*pfFoundMatchingNestedBehavior))
           {
              break; // (done, since only one match is needed)
           }
        }
        pBehavClassCurrNode = pBehavClassCurrNode->next;
     }

  Return from GenerateAndTestForCausativeSituation()
```

**GenerateAndTest_ProcessOneForwardRule ()** has the following inputs:

- A pointer to the object instance candidate (e.g. an object instance representing the “councilmen”, or an object instance representing the “demonstrators”). The object instance data structure also contains a pointer to the object frame class from which it is derived so that object frame class information, such as structural parent class can be obtained.
- A behavior class that has been retrieved by a prior search process that provided one or more object frame classes and a verb-based expression. An example such behavior class is called “TalkerAdvocatesActionWithListenersWhoAnticipateSomething” – this behavior class was retrieved based on the verb “advocates” along with other criteria.
- The pronoun feature set data structure; this includes information about the other syntactic and semantic entities of the clause or phase wherein the pronoun is contained. E.g. for “because they advocated violence”, it includes “violence” as a syntactic direct object and as an object that fills the actee semantic role within that clause.

This routine first creates the temporary working memory sandbox (West) context. The output of this routine as shown below is the West context as it has been added to by the insertion of a major structural parent instance, a minor structural parent instance, and object instances within the structural parent instances. The object instances have had their state attributes set with values that will later get matched against attribute values of other object instances from the East sandbox context in order to determine if the candidate is the correct antecedent for the unresolved pronoun.

Note that the example rule shown here contains an object for the “Talker” – this is handled as a single talker even though it needs to be matched against a possible group of talkers (e.g. councilmen or demonstrators) because the singular/plural aspect is not relevant for the inference process (either “councilman” or “councilmen” will work). In contrast, the “listeners” are represented as a collection since it is necessary to represent the fact that there is a set of possible listeners; there is logic that determines that that set can include the councilmen, for the cases where the councilmen are not the talker.

```
GenerateAndTest_ProcessOneForwardRule ()

// Original NL sentence example:
//   "The city councilmen refused the demonstrators a permit because they advocated violence."
//
//      INPUT: Main forward behavior class: TalkerAdvocatesActionWithListenersWhoAnticipateSomething
//
//       ANTECEDENT: (not shown)
//           ...
//        CONSEQUENT:
//
//            PopulatedObjectClass "ConsequentActor" ( // Talker
//               <ObjectFrameClass ref = PersonObjectFrameClass />
//               <Attribute ref = CommunicatingState val = "CommunicatingCompleted" />
//            );
//            PopulatedObjectClass "ConsequentActee" ( // Representation-of-Action
//               <ObjectFrameClass ref = CommunicationUnitProposedActionObjectFrameClass />  // e.g. violence
//               <PassiveParticipant val = "true" />
//               <Attribute ref = PassiveIsCommunicatedState val = "Communicated" />
//            );
//            PopulatedObjectClass "ConsequentExtra" (     // Listener(s)
//               <ObjectFrameClass ref = PersonObjectFrameClass />
//               <Multiple val = "true" />  // Collection
//               <ExtraParticipant val = "true" />
//               <Attribute ref = CommunicationReceivedState val = "CommunicationReceived" />
//               <Attribute ref = UniqueIdentityAttributeType var = extra$ />
//            );
//            // reference to a nested rule: this represents that whoever is the listener will fear violence:
//            BehaviorClassReference (
//               <BehaviorClass ref = AnticipateHarmfulEventBehaviorClass />
//               <ParameterActor ref = PersonObjectFrameClass expr = extra$ />  // (reference to the listener(s))
//               <ParameterActee ref = CognitiveRepresentationOfHarmfulEvent />
//            );

//------------------------------------------------------------------------------------------------------------------
//   (WEST SIDE)
//------------------------------------------------------------------------------------------------------------------

 //------------------------------------------------------------------------------------------------------------------
 // Create a new temporary context along with a structural parent instance (“major”),
 //    - sets context fields, and inserts the structural parent instance into the context.
 //    - (by default use the first ordinal temporal attribute value of the structural parent class)
 //
 CreateSandboxContext()

 //------------------------------------------------------------------------------------------------------------------
 // Create object instances and set values for semantic roles:
 //    - create clone of the candidate object instance (pObjInstCandidate)  // e.g. “councilmen”
 //    - use the pronoun feature set to determine other object instances, e.g. “violence”
 //
 EstablishObjectInstances()
```

```
//--------------------------------------------------------------------------------------------------------------------
// Attach all object instances to the structural parent ("major") within the sandbox context:
//
AttachObjectInstancesToStructuralParentMajorAndInstantiate()

//--------------------------------------------------------------------------------------------------------------------
// Invoke the Main Inference Routine:
//
PerformForwardDirectedInferenceWithNestedBehavior()

// (the West context has now been populated with all information based on the application of
//  the TalkerAdvocatesActionWithListenersWhoAnticipateSomething rule and the nested rule
//  that it contains - AnticipateHarmfulEventBehaviorClass

//--------------------------------------------------------------------------------------------------------------------
//   (EAST SIDE)
//--------------------------------------------------------------------------------------------------------------------

  ProcessLaterTemporalSandboxContextAndPerformMatchingTest ()


Return from  GenerateAndTest_ProcessOneForwardRule ()
```

Refer to Hofford (2014 (b)) "The ROSS User's Guide and Reference Manual*", 15.6.2. Main Inference Routine: Application of Two Rules* for details on the functionality of **PerformForwardDirectedInferenceWithNestedBehavior()**. The inference involves an application of the main rule combined with a subsequent application of the nested rule in order to derive new information that represents that there is a set of listeners that anticipate/fear a harmful event (i.e. which includes violence).

**ProcessLaterTemporalSandboxContextAndPerformMatchingTest ()** utilizes the later temporal information that was provided by the original clause (e.g. "the city councilmen refused the demonstrators a permit"). (This clause has already been used by the engine: the semantics that it represents exist in the master instance model and the spanning info data structure contains links that point at the relevant object instances).

Using the example, the main tasks are to iteratively: 1) derive a nested rule, if one exists, from the main rule (e.g. main rule = RefusingSomethingDueToFearBehaviorClass), 2) apply the nested rule (e.g. AnticipateHarmfulEventBehaviorClass), and 3) determine if the generated information (from this East side context) matches the earlier generated information from the West side context.

```
ProcessLaterTemporalSandboxContextAndPerformMatchingTest (
                        // IN:
                         ObjectInstance *pObjInstCandidate,
                         SpanningInformation *pSpanningInformation, // IN/OUT
                         BehaviorClassListNode *pBehavClassHeadNodeForCandidate, // list
                        // OUT:
                         bool *pfFoundMatchingNestedBehavior,
                         BehaviorClass **ppBehaviorClassNested)

  //--------------------------------------------------------------------------------------------------------------------
  // Loop: try each behavior class: e.g.:
  //
  //     - RefusingSomethingDueToFearBehaviorClass
```

```
   //      - RefusingSomethingDueToScheduleConflictBehaviorClass

   BehaviorClassListNode *pBehavClassCurrNode = pBehavClassHeadNodeForCandidate;

   while (pBehavClassCurrNode != NULL)
   {
      //---------------------------------------------------------------------------------------
      // Generate new object instances and states:
      //
      //   Class: PersonObjectFrameClass (q$)           ->  Instance: AnticipatingHarmfulEventState = "Anticipating"
      //   Class: CognitiveRepresentationOfHarmfulEvent  ->  Instance: PassiveIsAnticipatedState = "Anticipated"
      //

      CreateSandboxContext ()  // "East" side context

      // Populate the context/put the object instances into the structural parent within the sandbox context:

      CloneObjectInstancesFromSpanningInfoObjectInstances() ;
      AttachObjectInstancesToStructuralParentAndInstantiate();

      // Get nested rule from the next behavior class from the input list:

      pBehaviorClassNestedReference = NULL;

      GetNestedBehaviorClassReferenceFromBehaviorClass (pBehavClassCurrNode->pBehaviorClass,
                                                        &pBehaviorClassNestedReference);

      //  E.g. Nested behavior within RefusingSomethingDueToFearBehaviorClass:
      //
      //     <BehaviorClass ref = AnticipateHarmfulEventBehaviorClass />
      //     <ParameterActor ref = PersonObjectFrameClass expr = q$ />
      //     <ParameterActee ref = CognitiveRepresentationOfHarmfulEvent />

      // APPLY: Nested Rule/BehaviorClass: AnticipateHarmfulEventBehaviorClass
      ApplyBehaviorClass (pInfopedia,
                          &activePointersForInstancesPOST,
                          pBehaviorClassNestedReference->GetReferenceBehaviorClass());

      // Check for match: information from West (earlier) versus East (later)

      iResult = MatchNewObjectInstanceStatesLatestPriorAgainstEarliestPost (pStructuralParentLATESTPRIOR,
                                                                           pStructuralParentEARLIESTPOST);
      if (iResult == E_SUCCESS)
      {
         *pfFoundMatchingNestedBehavior = true;
         *ppBehaviorClassNested = pBehaviorClassForwardRule;

         break;  // !FOUND!
      }

      pBehavClassCurrNode = pBehavClassCurrNode->next;

   }  // while (pBehavClassCurrNode != NULL)

Return from ProcessLaterTemporalSandboxContextAndPerformMatchingTest ()
```

The details of **MatchNewObjectInstanceStatesLatestPriorAgainstEarliestPost()** are not shown. This function matches object instances, based on criteria that includes their respective object frame classes, and attribute state values. A set of example values is shown here:

```
//==========================================================================
// WEST: StructuralParent contains:
//   Instance: PersonObjectFrameClass (extra$)       ->  Attr:AnticipatingHarmfulEventState = "Anticipating"
//   Instance: CognitiveRepresentationOfHarmfulEvent  ->  Attr:PassiveIsAnticipatedState = "Anticipated"
//==========================================================================
// EAST: StructuralParent contains:
//   Instance: PersonObjectFrameClass (q$)          ->  Attr:AnticipatingHarmfulEventState = "Anticipating"
//   Instance: CognitiveRepresentationOfHarmfulEvent  ->  Attr:PassiveIsAnticipatedState = "Anticipated"
//==========================================================================
```

When MatchNewObjectInstanceStatesLatestPriorAgainstEarliestPost() is called for the "demonstrators" as the object instance candidate, the earlier application of the nested behavior on the West side had generated an object instance that is a set of "listeners" that is exclusive of the demonstrators. MatchNewObjectInstanceStatesLatestPriorAgainstEarliestPost () searches the West context for each object instance in the structural parent instance; it finds a set of "listeners" (PersonObjectFrameClass (extra$), above) that has an attribute with attribute type = "AnticipatingHarmfulEventState" and attribute value = "Anticipating". When the object instance from the East side context's structural parent instance contains a similar such object instance, the test succeeds: several flags are set and E_SUCCESS is returned from the function. The caller functions will utilize the information that the object instance candidate (e.g. the demonstrators) was successfully used by the set of inference processes and the instance model-based matching routine in order to determine its validity as a candidate referent.

## 9. Applications: Winograd Schemas #8, #115, #1

This section describes solutions for Winograd schemas #8, #115, and #1. (The two variants for schema #1 – "feared violence" and "advocated violence" are described separately). The solutions are described in terms of how they address the schemas as general pronoun resolution use cases.

*(The trophy and suitcase schema solution is described in Appendix 1: Solution for "Trophy and Suitcase" Schema Using a Model of the Communicating Agent).*

### 9.1. Schema #8: "The man could not lift his son"

This use case involves a main meaning unit that contains declarative text about a past situation, followed by second meaning unit (clause) that describes something that occurred or was true at an earlier point in time. An earlier event or state may exist within an explanatory ("because") clause. The pronoun in the current meaning unit refers back to an antecedent in the main meaning unit. This use case includes the following sub-use cases:

- Dependent clause introduced by "because", "since", etc., where the semantics of the clause are of an explanatory nature.
- Dependent adverbial clause introduced by "after".

The pronoun resolution algorithm makes use of the behavior class that was used to generate instance model information; in doing so it needs a specification of whether or not to examine the object frame classes of the behavior class's *PriorStates* (rule antecedent) section or the behavior class's *PostStates* (rule consequent section). Therefore the caller function for the pronoun resolution algorithm contains logic that sets an enumerated value for PredicateExpressionTemporalOrderIndicator (to PredicateExpressionTemporalOrderIndicatorPreceding). This is passed to PronounResolutionGeneralMethod() via the pronoun feature set structure.

### 9.1.1. How to Handle Duplicate Classes and Actor/Actee Identification

The "person lifts person" schema builds on the basic resolution method but it also needs a feature that is shown in the following code from "NotLift_Weak_BehaviorClass": the *PassiveParticipant* flag.

The ontology and knowledge base features that were used for this schema include:

- Object frame classes:
    - Person class and several sub-classes, including "man" and "son".
- Behavior classes:
    - NotLift_Weak_BehaviorClass
    - NotLift_Heavy_BehaviorClass

A portion of NotLift_Weak_BehaviorClass is shown here in order to illustrate the use of the functional attribute type that has a value that is used to match the "weak" of "too weak". This also illustrates the use of the passive participant flag.

```
BehaviorClass "NotLift_Weak_BehaviorClass"
...

    PopulatedObjectClass "AntecedentActor"
    (
      <ObjectFrameClass ref = PersonObjectFrameClass />
      <BinderSourceFlag val = "true" />
      <DimensionSystem ref = RelativePosition />
      <Attribute ref = RelativeLocation var = a$ />
      <Attribute ref = RelativeTime var = t1$ />
      <Attribute ref = LiftingState val = "NotLifting" />
      <Attribute ref = FunctionalAttributeType1 val = "TooWeak" />
    );
```

```
PopulatedObjectClass "AntecedentActee"
(
   <ObjectFrameClass ref = PersonObjectFrameClass />
   <PassiveParticipant val = "true" />
   <DimensionSystem ref = RelativePosition />
   <Attribute ref = RelativeLocation expr = (a$+1) />
   <Attribute ref = RelativeTime expr = t1$ />
   <Attribute ref = PassiveIsLiftedState val = "NotLifted" />
);
```

(The appendix contains full listings for the classes that are used in processing this schema).

### 9.2. Schema #115: "Joe paid the detective"

The "person pays detective" schema builds on the basic resolution method but it also needs a feature that is shown in the following code from the "PayAfterReceivingBehaviorClass" behavior class: the feature is the nested behavior. *(note: by way of comparison, in the terminology of Discourse Representation Theory, this is similar to an event within a DRS).*

The ontology and knowledge base features that were used for this schema include:

- Object frame classes:
  - Person class and several sub-classes: detective and deliverable. (The concept of a "deliverable" is used here as a high-level abstraction that includes any of services, products, a report, etc. This is one of several possible ways to model the semantics of the input schema text "received the final report").
- Behavior classes:
  - ReceiveBehaviorClass
  - DeliverBehaviorClass
  - PayAfterReceivingBehaviorClass
  - PersonIsPaidAfterDeliveringBehaviorClass

The nested behavior class feature is illustrated here in that PayAfterReceivingBehaviorClass contains a nested behavior class reference that refers to a separate behavior class called "ReceiveBehaviorClass" (full details are in the appendix).

```
BehaviorClass "PayAfterReceivingBehaviorClass"
(
…
   PriorStates
   (
      PopulatedObjectClass "AntecedentActor"
      (
         <ObjectFrameClass ref = PersonObjectFrameClass />
         <BinderSourceFlag val = "true" />
         <DimensionSystem ref = RelativePosition />
         <Attribute ref = RelativeLocation var = a$ />
         <Attribute ref = RelativeTime var = t1$ />
         <Attribute ref = PayingState val = "NotPaying" />
```

```
      <Attribute ref = UniqueIdentityAttributeType var = q$ />  // (identity)
   );
   BehaviorClassReference
   (
      <BehaviorClass ref = ReceiveBehaviorClass />  //  -->> DEFINED-BEHAVIOR-CLASS
      <ParameterActor ref = PersonObjectFrameClass expr = q$ /> // (identity)
      <ParameterActee ref = DeliverableObjectObjectFrameClass />
      <ParameterExtra ref = PersonObjectFrameClass />
   );
   PopulatedObjectClass "AntecedentActee"
   (
      <ObjectFrameClass ref = PersonObjectFrameClass />
      <PassiveParticipant val = "true" />
      <DimensionSystem ref = RelativePosition />
      <Attribute ref = RelativeLocation expr = (a$+1) />
      <Attribute ref = RelativeTime expr = t1$ />
      <Attribute ref = PassiveIsPayedState val = "NotPayed" />
   );
);
```

### 9.3. Schema #1/Variant #1: “City councilmen refused … feared violence”

The sentence for this schema is: “The city councilmen refused the demonstrators a permit because they *feared* violence.”. Resolution of the difficult pronoun for this variant of this schema uses the general pronoun resolution method; the embedded inference process is not needed since the instance model contains sufficient information to make the determination. The main ROSS rule that is used in the processing of the input sentence is the following. (This shows only key parts of the rule).

```
//------------------------------------------------------------------------------
//
//  BehaviorClass: "RefusingSomethingDueToFearBehaviorClass"
//
//    "If a person(s) anticipates a harmful event
//      then he/she/they will not grant a thing that was requested (e.g. a permit request)."
//
//------------------------------------------------------------------------------
//
BehaviorClass "RefusingSomethingDueToFearBehaviorClass"
(
   <CausalRule val = "true" />
   <BridgeObjectFrameClass ref = BehavioralStructuralParentClass />

   Dictionary ( English
   (
      { "refuse", "refused", "refused", "refuses", "refusing" }
   ););

   PriorStates  // (antecedent)
   (
      PopulatedObjectClass "AntecedentActor"
      (
         <ObjectFrameClass ref = PersonObjectFrameClass />  // e.g. government official(s)
         <Attribute ref = RefusingState val = "NotRefusing" />
```

```
            <Attribute ref = UniqueIdentityAttributeType var = q$ />
        );
        BehaviorClassReference  // ("if a person anticipates a harmful event")
        (
            // <Probability expr = 0.9 />  // (not used here)
            <BehaviorClass ref = AnticipateHarmfulEventBehaviorClass />  // --> defined behavior class
            <ParameterActor ref = PersonObjectFrameClass expr = q$ /> // (refers to the actor)
            <ParameterActee ref = CognitiveRepresentationOfHarmfulEvent />
        );

        // Actee (not shown) e.g. a person class from which demonstrators inherits

        // Extra (not shown)  e.g. a class from which permit inherits

    PostStates  // (consequent)
    (
        PopulatedObjectClass "ConsequentActor"
        (
            <ObjectFrameClass ref = PersonObjectFrameClass />  // e.g. government official(s)
            <Attribute ref = RefusingState val = "Refusing" />
        );

        // (others not shown)
    );
);
```

The nested behavior class is called "AnticipateHarmfulEventBehaviorClass" (not shown here).

A call stack for the processing of this schema is shown below *(note: this uses the meaning unit rather than the predicate expression)*. The call stack shows how the main meaning unit is processed first, by ProcessMeaningUnitIndicative() – this processing generates instance model object instances which constitute the possible referents for the pronoun that is subsequently processed. The spanning information structure points to the newly-created object instances in the instance model. The processing of the current meaning unit ("because they feared violence") occurs in a subsequent call to ProcessMeaningUnitIndicative(). In this case, the pronoun that needs resolution is in the syntactic subject, thus there are two intervening calls for handling the subject, leading to the call to EntityResolutionRoutine(). This function calls ProcessPronoun(), passing in a fully populated PronounFeatureSet data structure. Some fields of the PronounFeatureSet are used to direct the call (since the PronounFeatureSet designates the current meaning unit as a "because clause", PronounResolutionExplanatoryClauseForCognitiveExplanation() is invoked.

PronounResolutionWorkerRoutine() is the main anaphora resolution routine. It invokes ExploratorySearchUsingSpanningInfoStack(), a high-level driver function for the "exploratory searches" that will take place. This function loops through all spanning info's in the stack. ExploratorySearchUsingOneSpanningInfo() invokes TestOneCandidateObjectInstance() for each of the actor object instances, actee object instances and extra object instances in the spanning information data structure. TestOneCandidateObjectInstance() again looks at the PronounFeatureSet: if an adjective is available it attempts to match it as the causal feature. In this case a verb is available ("refused"), and thus

TryToMatchVerbCausalFeatureAgainstSpanningInfoObjectInstance() is invoked. Some details of TryToMatchVerbCausalFeatureAgainstSpanningInfoObjectInstanceSingleBehaviorClass() are not described here: the mechanism involves a set of preparatory steps leading up to the call to SearchUsingNestedBehaviorForMatchingCausalFeature(). The lowest-level function in this call stack is SearchUsingNestedBehaviorForMatchingCausalFeature(). It does these tasks:

- searches the behavior classes that are associated with the object instance candidate's class (e.g. CityCouncilmanObjectFrameClass) for one that matches the verb ("refused"). This retrieves a behavior class pointer.
- compares the nested behavior class that was supplied as a parameter to the behavior class pointer that was just retrieved – if they match, the object instance candidate is flagged as the pronoun antecedent and control returns up to PronounResolutionWorkerRoutine() for a set of follow-up tasks that include adding new information to the instance model.

Call Stack:

```
EngineDriver()
 ProcessCommunicationUnitList()
  ProcessMeaningUnitList()
   ProcessMeaningUnit()
    ProcessMeaningUnitIndicative()  // "The city councilmen refused the demonstrators a permit"
     ProcessPredicatePhrase()
      ProcessMeaningUnit()
       ProcessMeaningUnitIndicative() // "because they feared violence"
        ProcessSubjectPhrase()
         ProcessSubjectPhraseNounPhrase()
          EntityResolutionRoutine()
           ProcessPronoun()   // resolve "it"
            PronounResolutionExplanatoryClauseForCognitiveExplanation()  // handle a "because" clause
             PronounResolutionWorkerRoutine()
              ExploratorySearchUsingSpanningInfoStack()   // loop: search each spanning info in the stack
               ExploratorySearchUsingOneSpanningInfo()  // invoke the tests
                TestOneCandidateObjectInstance()  // branch based on available keyword
                                                   // ("refused") in PronounFeatureSet
                 TryToMatchVerbCausalFeatureAgainstSpanningInfoObjectInstance()  // loop for all
                                                                                 // behavior classes
                  TryToMatchVerbCausalFeatureAgainstSpanningInfoObjectInstanceSingleBehaviorClass()
                   SearchUsingNestedBehaviorForMatchingCausalFeature()
```

The present method can also utilize probability values that are associated with nested behaviors in order to select the most appropriate behavior class from among several. Refer to Hofford (2014 (b)) "The ROSS User's Guide and Reference Manual", Appendix: Star Classes for the Solution for Winograd Schema #1, 5.3.1. RefusingSomethingDueToFearOnPartOfRequestorBehaviorClass for full details.

### 9.4. Schema #1/Variant #2: “City councilmen refused … advocated violence”

The sentence for this schema is: “The city councilmen refused the demonstrators a permit because they *advocated* violence.”. Refer to section **8. Embedded Commonsense Reasoning Method** for details of this logic.

## 10. Conclusion: Test Results for Schemas

The method has been fully implemented in a working system that processes sentences, creates instance models, and then answers relevant questions based on its internal knowledge[7]. The Comprehendor system is also usable via a RESTful API server.

The following were derived from or directly adopted from the Winograd Schema Challenge schemas. (A minor change to one of the original schemas is noted).

(Note: the API call results show the antecedents in parenthesis after the pronoun (e.g. “it(trophy)”). The method is capable of determining the noun phrase, not just the noun head word and this will be addressed in a future version so that the system will generate the phrase, e.g. “it(*the trophy*) and “it(*the brown suitcase*)”).

### 10.1. Trophy and Suitcase Schema

#### 10.1.1. Original Winograd Schema

The trophy doesn't fit into the brown suitcase because it's too [small/large]. What is too [small/large]?

Answers: The suitcase/the trophy.

#### 10.1.2. Test Results

The test results were as follows.

>The trophy doesn’t fit in the brown suitcase because it’s too big.

>What is too big?

>The trophy is too big.

>The trophy doesn’t fit in the brown suitcase because it’s too small.

>What is too small?

>The suitcase is too small.

[7] Information about a demo for the working system is available on the author’s web site (Hofford (2014 (c)).

The same test can be run using Comprehendor as a RESTful API server. (This test and the following tests use the cURL tool to submit requests to the Comprehendor server). The responses are shown in italics. (Note: Comprehendor expands contractions, e.g. “doesn’t” -> “does not”; thus the resulting output sentence shows the un-contracted form of contractions that exist in the input sentence).

C:\ cURL>curl --data "Task=DisambiguateSentences&InputText=The trophy doesn’t fit in the brown suitcase because it’s too big." http://192.168.1.3/ServerMethod.NLUTask

*The trophy does not fit in the brown suitcase because it(trophy) is too big .*

C:\ cURL>curl --data "Task=DisambiguateSentences&InputText=The trophy doesn’t fit in the brown suitcase because it’s too small." http://192.168.1.3/ServerMethod.NLUTask

*The trophy does not fit in the brown suitcase because it(suitcase) is too small .*

## 10.2. Person Lifts Person Schema

### 10.2.1. Original Schema

The man couldn't lift his son because he was so [weak/heavy]. Who was [weak/heavy]?

Answers: The man/the son.

### 10.2.2. Test Results

The test results were as follows. (Changes: “couldn’t” changed to “didn’t”; “so” to “too”).

C:\ cURL>curl --data "Task=DisambiguateSentences&InputText=The man didn’t lift his son because he was too weak." http://192.168.1.3/ServerMethod.NLUTask

*The man did not lift his son because he(man) was too weak .*

C:\ cURL>curl --data "Task=DisambiguateSentences&InputText=The man didn’t lift his son because he was too heavy." http://192.168.1.3/ServerMethod.NLUTask

*The man did not lift his son because he(son) was too heavy .*

## 10.3. Person Pays Person Schema

### 10.3.1. Original Schema

Joe paid the detective after he [received/delivered] the final report on the case. Who [received/delivered] the final report?

Answers: Joe/the detective.

### 10.3.2. Test Results

The test results were as follows.

C:\cURL>curl --data "Task=DisambiguateSentences&InputText=Joe paid the detective after he received the final report on the case." http://192.168.1.3/ServerSideTask.NLUTask

*Joe paid the detective after he(Joe) received the final report on the case .*

C:\cURL>curl --data "Task=DisambiguateSentences&InputText=Joe paid the detective after he delivered the final report on the case." http://192.168.1.3/ServerSideTask.NLUTask

*Joe paid the detective after he(detective) delivered the final report on the case .*

## 10.4. Councilmen and Demonstrators Schema

### 10.4.1. Original Schema

The city councilmen refused the demonstrators a permit because they [feared /advocated] violence. Who [feared/advocated] violence?

Answers: The councilmen/the demonstrators.

### 10.4.2. Test Results

The test results were as follows.

C:\cURL >curl --data "Task=DisambiguateSentences&InputText=The city councilmen refused the demonstrators a permit because they feared violence." http://192.168.1.2/ServerMethod.NLUTask

*The city councilmen refused the demonstrators a permit because they(councilmen) feared violence .*

C:\\cURL>curl --data "Task=DisambiguateSentences&InputText=The city councilmen refused the demonstrators a permit because they advocated violence." http://192.168.1.2/ServerMethod.NLUTask

*The city councilmen refused the demonstrators a permit because they(demonstrators) advocated violence .*

## Appendix 1: Solution for "Trophy and Suitcase" Schema Using a Model of the Communicating Agent

The original anaphora resolution method, as developed by the author and applied to solve the trophy and suitcase schema, involved a paradigm that models the process of communication itself. This process of communication involves the following elements; each of these is represented by object instances in a ROSS instance model. The object instances are based on ontology classes, as follows:

- Intelligent/Communicative Agents:
  - The *communicating* agent or agents (also referred to as the *talker*).
  - The *listening* agent (the NLU system itself, also referred to as the *listener*)
- The information that is communicated
- Cognitive entities that belong to the communicating agent:
  - Beliefs and knowledge about facts and about behaviors of objects in the agent's environment
  - The processes of cognition – reasoning on the part of the agent
  - The process of communicating, i.e. conveying information to a listener

### 1. Rationale for Modeling the Communicating Agent

The "communicating agent paradigm" is useful for NLU and anaphora resolution situations that include the following:

- (NLU) Two-way or multi-way dialog or written communication (not covered here)
- (anaphora resolution) Sentences that involve ambiguity where the specific beliefs of the communicating agent are not known by the listener. An example would involve the following sentence:

"The bat did not hit the baseball because it moved too fast."

Here the pronoun *it* refers to either the bat or the baseball. It is possible to imagine two situations where this might be spoken or conveyed:

- By a batter who has just swing and missed a fast pitch: this person explains the cause of his or her not hitting the ball from the perspective of a cognitive rule that describes bats and baseballs, wherein the causative behavioral feature of interest is pitches (balls) that are so fast that they are missed.
- By a batting coach who is coaching a rookie batter who tends to swing too fast. The batting coach explains the cause from the perspective of a cognitive rule that involves a causative behavioral feature: missed pitches caused by a batter who swings too quickly.

Despite a possible lack of plausibility of this particular example, it illustrates that some instances of natural language understanding can benefit from a model that takes into consideration the probability that the talker has a particular set of beliefs constituting the causative aspects of external phenomena.

The probability aspects are not addressed here, but the mechanisms involved in modeling a communicating agent, communicated information, and cognitive entities are explained.

## 2. Overview

Two distinct aspects are involved for the sentence "The trophy didn't fit in the suitcase because it was too big"[8]. The first of these is the process of communication on the part of an intelligent agent, including the abstract cognition (mental) entities that exist in the mind/brain of the intelligent agent. The second of these involves a representation of the actual, or external situation that the intelligent agent describes – for this schema it is modeled as a physical process of attempting to fit an object (trophy) into another object (suitcase).

The trophy and suitcase example sentences associate a behavior (described by the verb phrase "does not fit") with an attributive state that describes a causal feature ("bigness" or "smallness").

## 3. Ontology

An overview is described here: the appendix contains full listings for some of these classes.

### 3.1. Object Frame Classes

These include the following:

- higher-level classes:
    - a special class for a structural parent object that is used to construct a 4D frame of reference for a situation (EverydayObjectStructuralParentClass)
    - a class of objects that are capable of being instantiated in an instance of a EverydayObjectStructuralParentClass (EverydayObjectFrameClass)
    - a class of objects that can fit into a container (EnclosableObjectFrameClass)
    - a class of container objects (ContainerObjectFrameClass)
    - a class of common objects that provides other attribute types such as color (CommonObjectFrameClass)
- a trophy class that inherits properties from the EverydayObjectFrameClass, the EnclosableObjectFrameClass, and the CommonObjectFrameClass.

[8] For purposes of analysis, the sentence that is used in the remainder of this section refers to the trophy and suitcase in past tense ("didn't fit", versus "doesn't fit").

- a suitcase class that inherits properties from the EverydayObjectFrameClass, the ContainerObjectFrameClass, and the CommonObjectFrameClass.
- human intelligent agent class: it models an agent that performs cognition, and communication – i.e. this is the person that communicates (spoken or written) the sentences of the schema (IntelligentAgentObjectFrameClass)
- a mental conceptual representation (referred to as an "image", although it is not necessarily pictorial) of a static or process-wise situation, e.g. an image of the process of a person attempting to fit a trophy into a suitcase (CognitiveImageForSituationObjectFrameClass)
- a mental conceptual representation of the intelligent agent's cognitive representation of the causal explanation for the situation (CognitiveExplanationObjectFrameClass)
- a higher-level more generic "information" class from which the cognitive causal explanation class gets most of its properties: (RepresentationOfCausalExplanationObjectFrameClass)
- the information items: the spoken or written forms of the sentence and its constituent parts (CommunicationUnitSentenceObjectFrameClass, CommunicationFragmentMeaningUnitObjectFrameClass, CommunicationFragmentWordObjectFrameClass)

### 3.2. Behavior Classes

Internal knowledge base behavior class definitions are used as specifications of *causality*. Definitions exist for each of the following:

- the "fitting" process or behavior (FitsBehaviorClass)
- the "not fitting" behavior (NotFit_Big_BehaviorClass, NotFit_Small_BehaviorClass)

## 4. Instance Model

The internal instance model is generated from the main sentence in several stages. ("The trophy doesn't fit in the brown suitcase because it's too big."). It consists of a *main/overall model* that contains an *embedded model*:

- The *main/overall instance model* represents the communicating agent (the talker) and the process of communicating information. This main instance model contains an embedded instance model that represents the "fitting" action of the actual situation. The main instance model also involves representation of a cognitive process (on the part of the talker) involving reasoning about the causal aspects of the embedded instance model.

- The *embedded instance model* represents the *actual situation*: it involves the objects – trophy, suitcase, person, and an instance of the behavior “to fit” as a process that occurs along a timeline (it is a 4D representation of the situation). It includes:
    - *attachment* of a *structural parent instance* that is based on a structural parent object frame class (EverydayObjectStructuralParentClass).
    - use of a *timeline* with time points such as “T01”, “T02”.
    - *object instances* with *attributes*:  suitcase, trophy and person instances. An example attribute for the trophy is one that represents “not fitted into”, at T01.
    - behavior instances are implicitly implemented along the timeline. This involves specification of the suitcase and the trophy in an initial state, specification of the next state involving the action of moving the trophy towards/into the suitcase, and a final state where the trophy comes to rest outside of the suitcase.
    - Once the pronoun is resolved, an attribute for “too big” or “too small” is added to the embedded instance model as an attribute of either the trophy or suitcase depending on which object has been determined as the pronoun antecedent.

## 5. Diagram of Overall Situation

The overall situation involves a process wherein an intelligent agent (the talker) communicates two clauses within a single sentence. (The first clause is "The trophy did not fit in the suitcase"; the second clause is "because it was too big"). **Figure 2** shows the cognitive and communicative aspects of the situation (note that the listener agent is not shown).

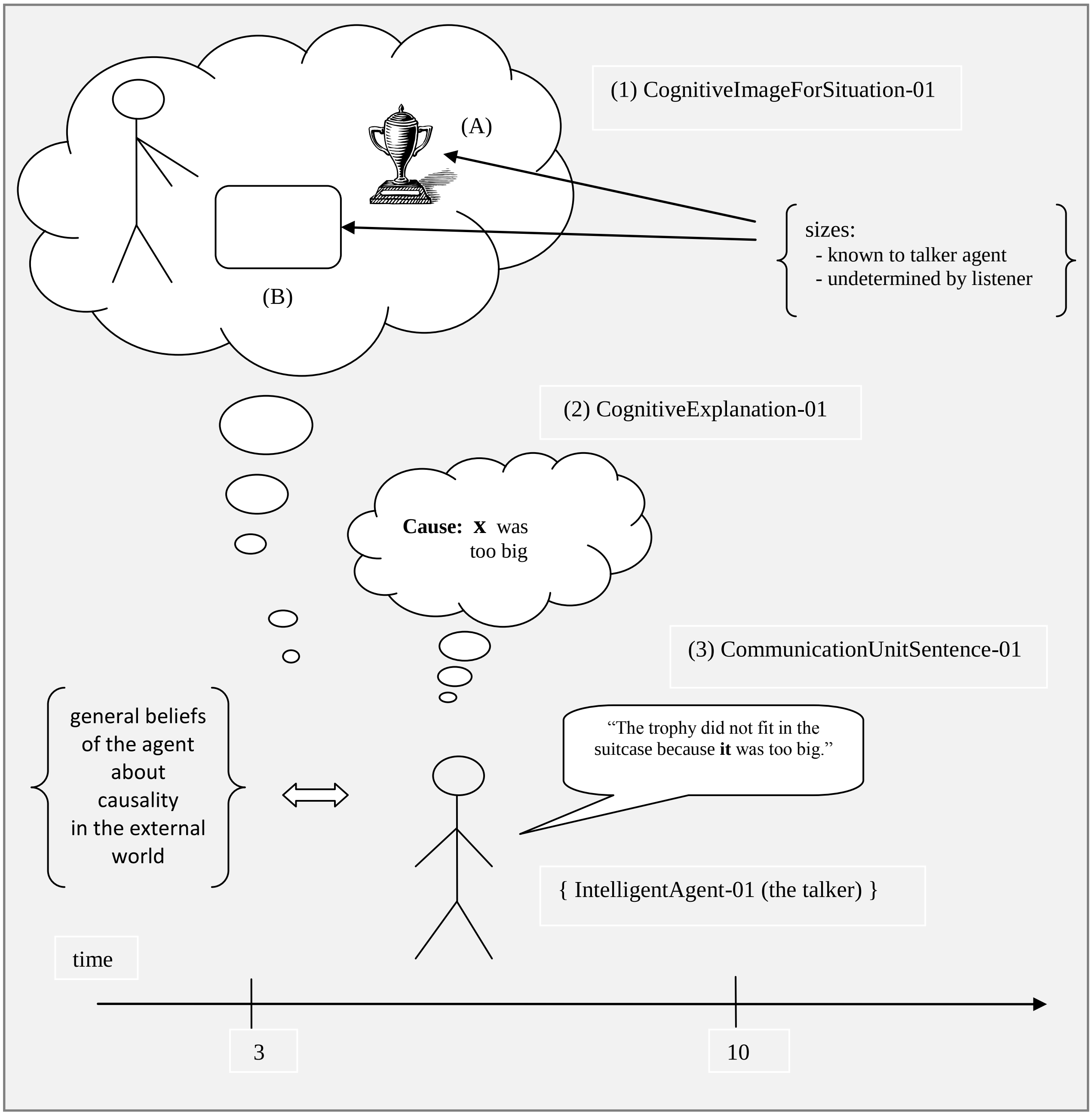

*Figure 2: Visualization of overall situation, for main instance model*

The semantic engine generates an overall instance model that corresponds to the diagram of figure 2. The overall instance model contains object instances, as follows. The intelligent agent is the talker, labeled as IntelligentAgent-01. The talker agent has a set of general beliefs (about how things work in the physical world) that is represented in brackets to the left of the this agent. An ActualPastSituation exists and occurred earlier, but it is only represented indirectly within the bubble in the upper left: this involves a trophy, a person, a suitcase and a "fitting attempt" action. (1) CognitiveImageForSituation-01 – this represents the intelligent agent's cognitive representation of the actual past situation. (2) CognitiveExplanation-01: this is what the intelligent agent believes about the cause(s) involved in the specific actual situation. (3) CommunicationUnitSentence-01 is an object instance that represents the sentence communicated by IntelligentAgent-01. CommunicationUnitSentence-01 consists of CommunicationFragmentMeaningUnit-01(not labeled – "The trophy did not fit in the suitcase") and CommunicationFragmentMeaningUnit-02 (not labeled – "because it was too big").

The timeline has two important time points. The timeline numbers have been selected only for illustration purposes and are intended to represent a typical scenario. (The working prototype uses an enumerated type consisting of timeline values such as "T01", "T02", etc.). At t=3 seconds, IntelligentAgent-01 forms the mental image as shown, and also at t=3 the intelligent agent forms a cognitive explanation of the past situation. At t = 10 the agent communicates by speaking "The trophy did not fit in the suitcase because it was too big".

## 6. Semantic Engine Tasks

The entity resolution and pronoun resolution tasks are described here. Underlying the pronoun resolution process is an assumption of a shared set of beliefs - between the talker agent and the listener agent - about "how things work". I.e. the listener agent builds a model of what the talker agent is thinking with respect to the talker's explanation of the cause of the trophy not fitting in the suitcase. The listener uses this model to reason about the possible meanings of the unresolved pronouns that were communicated to it.

### 6.1. Entity Resolution/Class Selection for Common Nouns and Verbs

The first task involves the selection of appropriate classes from the internal knowledge base that correspond to each of the common nouns and to the main verb phrase (representing "to *not* fit"). Although a ROSS knowledge base may contain classes that map the word "trophy" and "suitcase" to any of a number of classes, for the purpose of simplifying this example the following mappings from common words to knowledge base classes have been used:

- "trophy" – TrophyObjectFrameClass (inherits from EverydayObjectFrameClass, CommonObjectFrameClass and EnclosableObjectFrameClass) { an ordinary object with varying size, shape, color, composition, etc. }

- “suitcase” – SuitCaseObjectFrameClass (inherits from EverydayObjectFrameClass, CommonObjectFrameClass and from ContainerObjectFrameClass) { an ordinary object with varying size, shape, color, composition, etc. }
- “not fit” – NotFit_Big_BehaviorClass, NotFit_Small_BehaviorClass { behavior classes that are associated with both EnclosableObjectFrameClass and ContainerObjectFrameClass }

## 6.2. Pronoun Resolution

The pronoun resolution task (for “it” within the second clause) draws inferences about the undetermined entities, mental concepts, or words (each is represented by *x*).

- x within the *actual past situation* – e.g. which physical object was too big?
- x within the talker’s *cognitive image* of the past situation
- x within the talker’s *cognitive explanation* of the *causality* of the situation - i.e. within CognitiveExplanation -01 which item is associated with the (causal) feature that has a causative effect on the “not fitting” behavior?
- x within the natural language text: i.e. what is the pronoun antecedent within the first clause (“trophy” or “suitcase”)?

The output of the resolution process is a determination of the unknown facts, and is contained in the instance model for the overall situation that involves detail that includes the resolution of the pronoun.

The following is a form of pseudo-code using first-order logic to show the specifications of the rules that are used to support the inference. *(Since the logic within the antecedents for each of the three rules is similar, for the last two rules only the logic of the consequent is shown).* Each of the three rules contains several main groups of logical expressions in the antecedent:

- Expressions that specify the actual/external situation (e.g. the situation involving an instance of a trophy not fitting into a suitcase).
- Expressions that specify the natural language text itself (the main organizing predicate is “CommunicationUnitSentence”).
- Expressions that specify shared, or generally-known commonsense cognitive knowledge, e.g. about containers and things that can fit into containers (these expressions represent two ROSS behavior classes – one for “not fitting due to enclosable being too big”, and another for “not fitting due to container being too small”).
- Expressions that specify an instance of a CognitiveExplanationObjectFrameClass.

**Rule 1**: This rule uses an abstraction centering around an instance of a CognitiveExplanationObjectFrameClass. (The CognitiveExplanationObjectFrameClass inherits its properties from a RepresentationOfCausalExplanationObjectFrameClass class, which is actually used). This rule resolves the referent for a representational entity (the unknown entity) that exists within an instance of RepresentationOfCausalExplanationObjectFrameClass, referred to using the variable name "unknown-entity". This rule describes the logic that is used to figure out what the intelligent agent was thinking when he/she said "because it was too big". (Note that this rule example uses "ContainerClass" as an equivalent for the Infopedia class called "ContainerObjectFrameClass"). *(Note: logic for the "too small" case not shown).*

Rule 1: for "x was too big"

∀ unknown-entity**: // (Rule Antecedent)**

// Variables that represent specific attribute values:

∃attval1: CausalFeatureAttributeValue(attval1) ∧ StringValue(attval1, "TooBig")
∃attval2: CausalFeatureAttributeValue(attval2) ∧ StringValue(attval2, "TooSmall")

// The actual past situation. E.g. this models the failed "fitting attempt" instance, where *something* was too big:

∃sit: Situation(sit) ∧ // (@ t = n-2)
∃entity1: IsRepresentedByClassName(entity1, entity-class-name1) ∧ PartOf(entity1,sit) ∧ // e.g. TrophyClass
∃entity2: IsRepresentedByClassName (entity2, entity-class-name2) ∧ PartOf(entity2,sit) ∧ // e.g. SuitcaseClass
∃action1: AttemptToFitEnclosableIntoContainer(action1) ∧ PartOf(action1,sit) ∧

NotFittedInsideContainer(entity1) ∧ // Att-type = PassiveIsFittedInsideContainerState
NotIsFittedInto(entity2) ∧ // Att-type = PassiveIsFittedIntoState

( CausalFeature(entity1,atttype-name, attval1)
∨ // disjunction
CausalFeature(entity2,atttype-name, attval1) )

∧

// The input natural language text (the sentence and its constituent parts)

∃s: CommunicationUnitSentence (s) ∧ // the input sentence // (@ t = n)
∃m1,m2: // the two clauses of the sentence
MeaningUnit (m1) ∧ PartOf(m1,s) ∧ // clause that is a description of a past situation
∃subj: ReferentPhraseSubject(subj,entity1) ∧ PartOf(subj,m1) ∧ // e.g. enclosable object
∃adv: VerbModifierWord(adv) ∧ PartOf(adv,m1) ∧ // e.g. negation
∃verb: WordBehavior(verb) ∧ PartOf(verb,m1) ∧ // e.g. "fitting" behavior
∃dobj: ReferentPhraseDirObject(dobj,entity2) ∧ PartOf(dobj,m1) ∧ // e.g. container object

MeaningUnit (m2) ∧ PartOf(m2,s) ∧ // clause that is a causal explanation for the situation
∃adv2: CauseExplanationIntroducerWord(adv2) ∧ PartOf(adv2,m2) ∧ // e.g. "because"
∃pron: PronounWord(pron) ∧ PartOf(pron,m2) ∧ // the unresolved "variable"
∃verbtobe: AuxiliaryVerbToBeWord(verbtobe) ∧ PartOf(verbtobe,m2) ∧ // e.g. "was"
∃:caufeat: ReferentPhraseCausalFeatureAttValue(caufeat, attval1) ∧ PartOf(caufeat,m2) // "too big"
∧

// (Meta) Common/shared cognitive knowledge about behaviors (these exist in the ontology)

// Higher-level Entity classes:
∃enclosable: EnclosableObjectFrameClass(enclosable) ∧
∃container: ContainerObjectFrameClass(container) ∧

// Inherited Entity classes:
∃trophy: InheritsPropertiesFrom(enclosable) Λ
∃suitcase: InheritsPropertiesFrom (container) Λ

// (1) Behavior class for "an enclosable does not fit in a container if the enclosable is too big"
//
∃b1: CognitiveRepresentationOfBehaviorClass(b1) Λ // (@ t = any)
∃b1a: CogReprAntecedent(b1a) Λ PartOf(b1a, b1) Λ
∃reprentity1: Represents(repentity1, enclosable) Λ PartOf(repentity1,b1a) Λ
CausalFeature(reprentity1,atttype-name, attval1)
∃reprentity2: Represents(repentity2, container) Λ PartOf(repentity2,b1a) Λ
∃repaction: CogReprAction(repaction, action) Λ PartOf(repaction,b1) Λ
∃b1c: CogReprConsequent Λ PartOf(b1c, b1) Λ
*// (the following is shorthand for the Consequent result states)*
NotFittedInsideContainer(repentity1) Λ // Att-type = PassiveIsFittedInsideContainerState
NotIsFittedInto(repentity2) Λ // Att-type = PassiveIsFittedIntoState
Λ
// (2) Behavior class for "an enclosable does not fit in a container if the container is too small"
//
∃b2: CognitiveRepresentationOfBehaviorClass(b2) Λ // (@ t = any)
∃b2a: CogReprAntecedent(b2a) Λ PartOf(b2a,b2) Λ
∃reprentity1: Represents(repentity1, enclosable) Λ PartOf(repentity1,b2a) Λ
∃reprentity2: Represents(repentity2, container) Λ PartOf(repentity2,b2a) Λ
CausalFeature(reprentity2,atttype-name, attval2)
∃repaction: CogReprAction(repaction, action) Λ PartOf(repaction,b2) Λ
∃b2c: CogReprConsequent Λ PartOf(b2c, b2) Λ
*// (the following is shorthand for the Consequent result states)*
NotFittedInsideContainer(repentity1) Λ // Att-type = PassiveIsFittedInsideContainerState
NotIsFittedInto(repentity2) Λ // Att-type = PassiveIsFittedIntoState
Λ

// (Meta) The agent's cognitive explanation of the causal aspects of this situation:

∃ce: RepresentationOfCausalExplanationObjectFrameClass(ce) Λ // (@ t = n-1)
∃ce-cause: Repr-CauseEntity(cexplcause) Λ PartOf(ce-cause,ce) Λ
∃unknown-entity: Repr-AntecedentCausalAgent Λ PartOf(unknown-entity, ce-cause) Λ

( RepresentedClassName(unknown-entity, entity-class-name1) Λ // e.g. "TrophyClass"
V // disjunction
RepresentedClassName(unknown-entity, entity-class-name2)) Λ // e.g. "SuitcaseClass"

CausalFeatureAttributeTypeName(unknown-entity,atttype-name) Λ // e.g. "FunctionalSize"
CausalFeatureAttributeValue(unknown-entity, attval1) Λ // e.g. "TooBig"

**→ // (Rule Consequent) // (@ t = n)**

RepresentationalRelationship(unknown-entity, entity-class-name1)

The rule consequent expresses the fact that the unknown entity (within the mind of the cognitive agent) has a representational relationship with entity-class-name-1, which is an instance of the trophy class. What is not shown here (due to the complexities of specifying it with FOL), is the way in which the substitution takes place: the pronoun resolution algorithm uses the behavior class of the embedded situation (the "not fitting due to too big" behavior) in order to derive the higher class of the object (the enclosable object or the container object) that can affect the behavior result, which in this case is the enclosable object. It uses this class to determine which actual object is referred to based on the inheritance tree for the object frame class of each of the objects. *(Note: for purpose of ontology scalability, the use of probability values is viewed as a necessary requirement: given that both a trophy and a suitcase can be a container object (and each can be an enclosable object), the use of a probability value for the higher classes mechanism within the object frame class is envisioned. For instance, this would specify by implication (comparison of respective probability values) that it is more likely that a suitcase is a container than is a trophy).*

**Rule 2**: This rule associates the unresolved NL text pronoun (e.g. "it") with an entity in the represented, or external world. Given the unresolved pronoun within the explanatory ("because clause"), it resolves which *actual entity* in the external situation is referred to (e.g. the trophy object instance or the suitcase object instance). (The Comprehendor engine implements this rule in code subsequent to the resolution of Rule 1; it sets the appropriate attribute for the entity of the actual situation (FunctionalSize = "TooBig", or FunctionalSize = "TooSmall"). (note: "pron" designates the pronoun).

```
∀ pron:
   // (rule antecedent not shown)
→   // (Rule Consequent)   // (@ t = n)
  //  relationship of pronoun to actual situation entity:
      RepresentationalRelationshipWordToActualSituationEntity (pron, entity)
```

**Rule 3**: This rule centers around the natural language text. Given the unresolved pronoun within the explanatory ("because clause"), it resolves which *word* (anaphor antecedent) in the main clause the pronoun refers to (e.g. "trophy" or "suitcase").

```
∀ pron:
   // (rule antecedent not shown)
→   // (Rule Consequent)   // (@ t = n)
// relationship of pronoun to CommunicationFragmentWord:
      RepresentationalRelationshipCoreferent (pron, subj)
```

Note that in the actual system Rule 3 is not implemented, since the question answering system of the semantic engine is capable of searching the actual instance model that corresponds to the communicated sentence.

## Appendix 2: Ontology/Knowledge Base

The main Star language object frame classes and behavior classes that are used by the Comprehendor NLU system to process the schemas are shown in this appendix, with the exception of those for schema #1, which is contained in Hofford (2014 (b)) “The ROSS User’s Guide and Reference Manual”. The Star language definitions exist within several different Infopedia include files; these include files such as BasicDefinitions.h and PersonRelatedClass.h.

## 1. Ontology and KB for Schema: “Trophy and Suitcase”

Most of the middle and lower ontology classes that are needed for this schema were auto-generated from natural language input, using the Comprehendor Ontology Builder sub-system. The actual sentences are shown here:

Natural language input:

A container object is an everyday object.
An enclosable object is an everyday object that fits in a container object.
If an enclosable object is too big then it does not fit in the container object.
If a container object is too small then an enclosable object does not fit in it.
A trophy is an enclosable object.
A suitcase is a container object.

### 1.1. Supporting Definitions

Supporting definitions are described in the “ROSS User’s Guide and Reference Manual” (Hofford 2014 (b)).

### 1.2. Object Frame Classes

#### 1.2.1. Upper Ontology Classes

The following upper ontology classes/definitions are described in the “ROSS User’s Guide and Reference Manual” (Hofford 2014 (b)).

- Structural parent class: EverydayObjectStructuralParentClass
- Higher-level class: EverydayObjectFrameClass
- Structural parent class: BehavioralStructuralParentClass

#### 1.2.2. Middle Ontology Classes: Auto-Generated Classes

*(Note: comments were added by hand after completion of the auto-generation process).*

```
//------------------------------------------------------------------
//
//  ContainerObjectObjectFrameClass
//
//------------------------------------------------------------------
//
ObjectFrameClass "ContainerObjectObjectFrameClass"
(
   <StructureTrait val = "Compound"/>

   DictionaryPriorWord
   (
      <DictionaryWordIsNoun val = "true" />

      English
      (
         { "container",
           "containers" }
      );
   );

   Dictionary ( English
   (
      { "object",
        "objects" }
   ););

   HigherClasses ( { "EverydayObjectFrameClass" } );

   AttributeTypes
   (
      AttributeType "PassiveIsFittedState"  // or, “PassiveIsFittedIntoState”
      (
         <SuperType val = "Qualitative"/>

         <StateAttributeType val = "true" />

         "Values"
         (
            {
              "NotFitted",  // i.e. “NotFittedInto”  (not containing an object)
              "Fitted"         // i.e. “FittedInto” (containing an object)
            }
         );
      );

      AttributeType "FunctionalAttributeType1"
      (
         <SuperType val = "Qualitative"/>

         <StateAttributeType val = "true" />

         <OptionalCausalFeature val = "true" />

         "Values"
         (
            {
              "NotTooSmall",
```

```
              "TooSmall" : Dictionary ( English ( { "small" } ); );
            }
          );
        );
      );
    );

//------------------------------------------------------------------
//
//  EnclosableObjectObjectFrameClass  // e.g. a trophy, an apple
//
//------------------------------------------------------------------
//
ObjectFrameClass "EnclosableObjectObjectFrameClass"
(
   <StructureTrait val = "Compound"/>

   DictionaryPriorWord
   (
      English
      (
         { "enclosable",
           "enclosables" }
      );
   );

   Dictionary ( English
   (
      { "object",
        "objects" }
   ););

   HigherClasses ( { "EverydayObjectFrameClass" } );

   AttributeTypes
   (
      AttributeType "FittingState"
      (
         <SuperType val = "Qualitative"/>

         <StateAttributeType val = "true" />

         "Values"
         (
            {
              "NotFitting",  // e.g. not starting motion to fit into something
              "Fitting"        // e.g. in motion to fit into something
            }
         );
      );

      AttributeType "FunctionalAttributeType1"
      (
         <SuperType val = "Qualitative"/>

         <StateAttributeType val = "true" />

         <OptionalCausalFeature val = "true" />
```

```
          "Values"
          (
             {
               "NotTooBig",
               "TooBig" : Dictionary ( English ( { "big" } ); );
             }
          );
       );
    );
);

//---------------------------------------------------------------------------
//
//  "CommonObjectFrameClass"  // (not auto-generated)
//
//---------------------------------------------------------------------------
//
ObjectFrameClass "CommonObjectFrameClass"
(
    <StructureTrait val = "Compound"/>

    HigherClasses
    (
       {  "ObjectObjectFrameClass",  // (not shown in this document)
          "EverydayObjectFrameClass",
          "EarthBoundObjectFrameClass" }  // (not shown in this document)
    );

    StructuralParentClassesBase
    (
       {  "EverydayObjectStructuralParentClass" }
    );

    AttributeTypes
    (
       AttributeType "ExteriorColor"
       (
          <SuperType val = "Qualitative"/>

          "Values"
          (
             { "Black"   :  Dictionary ( English ( { "black" } ); ); ,
               "Silver"  :  Dictionary ( English ( { "silver" } ); ); ,
               "White"   :  Dictionary ( English ( { "white" } ); ); ,
             }
          );
       );

       // (one of many possible state attributes)
       AttributeType "StolenState"
       (
          <SuperType val = "Qualitative"/>
          <StateAttributeType val = "true"/>

          "Values"
          (
             { "NotStolen",
```

```
                "Stolen" }
          );
       );
    );

    // other attribute types here not shown

    DimensionSystems ();

    Structure ();

);  // "CommonObjectFrameClass"
```

### 1.2.3. Lower Ontology: Auto-Generated Classes

```
ObjectFrameClass "TrophyObjectFrameClass"
(
   <StructureTrait val = "Compound"/>

   Dictionary ( English
   (
      { "trophy",
        "trophys" }  // (bug in morphology analyzer: should be generated as “trophies”)
   ););

   HigherClasses ( { "EnclosableObjectObjectFrameClass" } );
);

ObjectFrameClass "SuitcaseObjectFrameClass"
(
   <StructureTrait val = "Compound"/>

   Dictionary ( English
   (
      { "suitcase",
        "suitcases" }
   ););

   HigherClasses
    (
       { "ContainerObjectObjectFrameClass",
         //TODO: add "CommonObjectFrameClass",  // (has color attribute type)
       }
    );
);
```

Note that the trophy and suitcase classes that are shown here only inherit properties from the enclosable object and container object classes, respectively. Since ROSS allows for multiple inheritance, other middle ontology classes can be added to the “HigherClasses” lists, e.g. “PropertyObjectFrameClass” (a class of objects that are owned as property).

### 1.3. Auto-Generated Behavior Classes

#### 1.3.1. Observations

The behavior classes shown here were also generated from the same sentences (above), copied here for clarity:

Natural language input:

An enclosable object is an everyday object that fits in a container object. // (positive case)
If an enclosable object is too big then it does not fit in the container object.
If a container object is too small then an enclosable object does not fit in it.

Note that the first behavior class below is a positive "fits" behavior class that is shown for comparision purposes (it is not used by the trophy and suitcase schema).

#### 1.3.2. Listing

```
BehaviorClass "FitsBehaviorClass"  // (positive case)
(
   <BridgeObjectFrameClass ref = BehavioralStructuralParentClass />

   Dictionary ( English
   (
      {
          "fit",     // (infinitive/base)
          "fitted",  // (simple past)
          "fitted",  // (past participle)
          "fits",    // (simple present, 3rd p.s.)
          "fitting"  // (present participle)
      }
   ););

   PriorStates
   (
      PopulatedObjectClass "AntecedentActor"
      (
         <ObjectFrameClass ref = EnclosableObjectObjectFrameClass />
         <BinderSourceFlag val = "true" />
         <DimensionSystem ref = RelativePosition />
         <Attribute ref = RelativeLocation var = a$ />
         <Attribute ref = RelativeTime var = t1$ />
         <Attribute ref = FittingState val = "NotFitting" />
      );
      PopulatedObjectClass "AntecedentActee"
      (
         <ObjectFrameClass ref = ContainerObjectObjectFrameClass />
         <PassiveParticipant val = "true" />
         <DimensionSystem ref = RelativePosition />
         <Attribute ref = RelativeLocation expr = (a$+1) />
         <Attribute ref = RelativeTime expr = t1$ />
```

```
      <Attribute ref = PassiveIsFittedState val = "NotFitted" />
    );
  );
  PostStates
  (
    PopulatedObjectClass "ConsequentActor"
    (
      <ObjectFrameClass ref = EnclosableObjectObjectFrameClass />
      <DimensionSystem ref = RelativePosition />
      <Attribute ref = RelativeLocation expr = (a$+1) />
      <Attribute ref = RelativeTime expr = (t1$+1) />
      <Attribute ref = FittingState val = "Fitting" />
    );
    PopulatedObjectClass "ConsequentActee"
    (
      <ObjectFrameClass ref = ContainerObjectObjectFrameClass />
      <PassiveParticipant val = "true" />
      <DimensionSystem ref = RelativePosition />
      <Attribute ref = RelativeLocation expr = (a$+1) />
      <Attribute ref = RelativeTime expr = (t1$+1) />
      <Attribute ref = PassiveIsFittedState val = "Fitted" />
    );
  );
);  // FitsBehaviorClass

BehaviorClass "NotFit_Big_BehaviorClass"
(
  <CausalRule val = "true" />
  <BridgeObjectFrameClass ref = BehavioralStructuralParentClass />
  <Negation val = "true" />

  Dictionary ( English
  (
    { "fit", "fit", "fitted", "fits", "fitting" }
  ););

  PriorStates
  (
    PopulatedObjectClass "AntecedentActor"
    (
      <ObjectFrameClass ref = EnclosableObjectObjectFrameClass />
      <BinderSourceFlag val = "true" />
      <DimensionSystem ref = RelativePosition />
      <Attribute ref = RelativeLocation var = a$ />
      <Attribute ref = RelativeTime var = t1$ />
      <Attribute ref = FittingState val = "NotFitting" />
      <Attribute ref = FunctionalAttributeType1 val = "TooBig" /> // (optional causal feature)
    );
    PopulatedObjectClass "AntecedentActee"
    (
      <ObjectFrameClass ref = ContainerObjectObjectFrameClass />

      <PassiveParticipant val = "true" />
      <DimensionSystem ref = RelativePosition />
      <Attribute ref = RelativeLocation expr = (a$+1) />
      <Attribute ref = RelativeTime expr = t1$ />
      <Attribute ref = PassiveIsFittedState val = "NotFitted" />
    );
```

```
    );
    PostStates
    (
      PopulatedObjectClass "ConsequentActor"
      (
        <ObjectFrameClass ref = EnclosableObjectObjectFrameClass />
        <DimensionSystem ref = RelativePosition />
        <Attribute ref = RelativeLocation expr = (a$+1) />
        <Attribute ref = RelativeTime expr = (t1$+1) />
        <Attribute ref = FittingState val = "Fitting" />
      );
      PopulatedObjectClass "ConsequentActee"
      (
        <ObjectFrameClass ref = ContainerObjectObjectFrameClass />
        <PassiveParticipant val = "true" />
        <DimensionSystem ref = RelativePosition />
        <Attribute ref = RelativeLocation expr = (a$+1) />
        <Attribute ref = RelativeTime expr = (t1$+1) />
        <Attribute ref = PassiveIsFittedState val = "Fitted" />
      );
    );
  );

  BehaviorClass "NotFit_Small_BehaviorClass"
  (
    <CausalRule val = "true" />
    <BridgeObjectFrameClass ref = BehavioralStructuralParentClass />
    <Negation val = "true" />

    Dictionary ( English
    (
      { "fit", "fit", "fitted", "fits", "fitting" }
    ););

    PriorStates
    (
      PopulatedObjectClass "AntecedentActor"
      (
        <ObjectFrameClass ref = EnclosableObjectObjectFrameClass />
        <BinderSourceFlag val = "true" />
        <DimensionSystem ref = RelativePosition />
        <Attribute ref = RelativeLocation var = a$ />
        <Attribute ref = RelativeTime var = t1$ />
        <Attribute ref = FittingState val = "NotFitting" />
      );
      PopulatedObjectClass "AntecedentActee"
      (
        <ObjectFrameClass ref = ContainerObjectObjectFrameClass />

        <PassiveParticipant val = "true" />
        <DimensionSystem ref = RelativePosition />
        <Attribute ref = RelativeLocation expr = (a$+1) />
        <Attribute ref = RelativeTime expr = t1$ />
        <Attribute ref = PassiveIsFittedState val = "NotFitted" />
        <Attribute ref = FunctionalAttributeType2 val = "TooSmall" /> // (optional causal feature)
      );
    );
    PostStates
```

```
  (
    PopulatedObjectClass "ConsequentActor"
    (
      <ObjectFrameClass ref = EnclosableObjectObjectFrameClass />
      <DimensionSystem ref = RelativePosition />
      <Attribute ref = RelativeLocation expr = (a$+1) />
      <Attribute ref = RelativeTime expr = (t1$+1) />
      <Attribute ref = FittingState val = "Fitting" />
    );
    PopulatedObjectClass "ConsequentActee"
    (
      <ObjectFrameClass ref = ContainerObjectObjectFrameClass />
      <PassiveParticipant val = "true" />
      <DimensionSystem ref = RelativePosition />
      <Attribute ref = RelativeLocation expr = (a$+1) />
      <Attribute ref = RelativeTime expr = (t1$+1) />
      <Attribute ref = PassiveIsFittedState val = "Fitted" />
    );
  );
);
```

## 2. Ontology and KB for Schema: "Person Lifts Person"

The following classes were auto-generated using the Comprehendor NLU system's Ontology Builder sub-system. The natural language input that was used is as follows. *(note: although the examples use "he" and "him", he/she and him/her can be used here interchangeably)*.

Natural language input:

If a person is too weak then he cannot lift another person.
If a person is too heavy then another person cannot lift him.

Since the ontology already contains a person class, the Comprehendor Ontology Builder sub-system only needed to generate lower ontology object frame class attributes and behavior classes. These classes are shown here. This demonstrates the Comprehendor Ontology Builder "partial class definition" feature, which allows for adding attributes or other information to a class that already exists in the ontology. For instance, the ontology contains PersonObjectFrameClass; the first definition below adds several attribute types to this class: "FunctionalAttributeType1", "LiftingState", and "PassiveIsLiftedState". A second definition adds another attribute type called "FunctionalAttributeType2".

### 2.1. Object Frame Classes

#### 2.1.1. Lower Ontology: Auto-Generated Class Information

```
ObjectFrameClass "PersonObjectFrameClass"
(
  <StructureTrait val = "Compound"/>

  AttributeTypes
  (
    AttributeType "FunctionalAttributeType1"
    (
      <SuperType val = "Qualitative"/>

      <StateAttributeType val = "true" />

      <OptionalCausalFeature val = "true" />

      "Values"
      (
        {
         "NotTooWeak",
         "TooWeak" : Dictionary ( English ( { "weak" } ); );
        }
      );
    );

    AttributeType "LiftingState"
    (
```

```
            <SuperType val = "Qualitative"/>

            <StateAttributeType val = "true" />

            "Values"
            (
                {
                  "NotLifting",
                  "Lifting"
                }
            );
        );

        AttributeType "PassiveIsLiftedState"
        (
            <SuperType val = "Qualitative"/>

            <StateAttributeType val = "true" />

            "Values"
            (
                {
                  "NotLifted",
                  "Lifted"
                }
            );
        );
    );

);

ObjectFrameClass "PersonObjectFrameClass"
(
    <StructureTrait val = "Compound"/>

    AttributeTypes
    (
        AttributeType "FunctionalAttributeType2"
        (
            <SuperType val = "Qualitative"/>

            <StateAttributeType val = "true" />

            <OptionalCausalFeature val = "true" />

            "Values"
            (
                {
                  "NotTooHeavy",
                  "TooHeavy" : Dictionary ( English ( { "heavy" } ); );
                }
            );
        );
    );
);
```

### 2.2. Behavior Classes

#### 2.2.1. Observations

The behavior classes shown here were also generated from the same two sentences (above), copied here for clarity.

Natural language input:

If a person is too weak then he cannot lift another person.
If a person is too heavy then another person cannot lift him.

#### 2.2.2. Listing

```
BehaviorClass "NotLift_Weak_BehaviorClass"
(
  <CausalRule val = "true" />

  <BridgeObjectFrameClass ref = BehavioralStructuralParentClass />

  <Negation val = "true" />

  Dictionary ( English
  (
    {
     "lift",
     "lifted",
     "lifted",
     "lifts",
     "lifting"
    }
  ););

  PriorStates
  (
    PopulatedObjectClass "AntecedentActor"
    (
      <ObjectFrameClass ref = PersonObjectFrameClass />
      <BinderSourceFlag val = "true" />
      <DimensionSystem ref = RelativePosition />
      <Attribute ref = RelativeLocation var = a$ />
      <Attribute ref = RelativeTime var = t1$ />
      <Attribute ref = LiftingState val = "NotLifting" />
      <Attribute ref = FunctionalAttributeType1 val = "TooWeak" />
    );
    PopulatedObjectClass "AntecedentActee"
    (
      <ObjectFrameClass ref = PersonObjectFrameClass />
      <PassiveParticipant val = "true" />
      <DimensionSystem ref = RelativePosition />
      <Attribute ref = RelativeLocation expr = (a$+1) />
      <Attribute ref = RelativeTime expr = t1$ />
```

```
        <Attribute ref = PassiveIsLiftedState val = "NotLifted" />
      );
   );
   PostStates
   (
      PopulatedObjectClass "ConsequentActor"
      (
         <ObjectFrameClass ref = PersonObjectFrameClass />
         <DimensionSystem ref = RelativePosition />
         <Attribute ref = RelativeLocation expr = (a$+1) />
         <Attribute ref = RelativeTime expr = (t1$+1) />
         <Attribute ref = LiftingState val = "Lifting" />
      );
      PopulatedObjectClass "ConsequentActee"
      (
         <ObjectFrameClass ref = PersonObjectFrameClass />
         <PassiveParticipant val = "true" />
         <DimensionSystem ref = RelativePosition />
         <Attribute ref = RelativeLocation expr = (a$+1) />
         <Attribute ref = RelativeTime expr = (t1$+1) />
         <Attribute ref = PassiveIsLiftedState val = "Lifted" />
      );
   );
);

BehaviorClass " NotLift_Heavy_BehaviorClass "
(
   <CausalRule val = "true" />

   <BridgeObjectFrameClass ref = BehavioralStructuralParentClass />

   <Negation val = "true" />

   Dictionary ( English
   (
      {
       "lift",
       "lifted",
       "lifted",
       "lifts",
       "lifting"
      }
   ););

   PriorStates
   (
      PopulatedObjectClass "AntecedentActor"
      (
         <ObjectFrameClass ref = PersonObjectFrameClass />
         <BinderSourceFlag val = "true" />
         <DimensionSystem ref = RelativePosition />
         <Attribute ref = RelativeLocation var = a$ />
         <Attribute ref = RelativeTime var = t1$ />
         <Attribute ref = LiftingState val = "NotLifting" />
      );
      PopulatedObjectClass "AntecedentActee"
      (
         <ObjectFrameClass ref = PersonObjectFrameClass />
```

```
            <PassiveParticipant val = "true" />
            <DimensionSystem ref = RelativePosition />
            <Attribute ref = RelativeLocation expr = (a$+1) />
            <Attribute ref = RelativeTime expr = t1$ />
            <Attribute ref = PassiveIsLiftedState val = "NotLifted" />
            <Attribute ref = FunctionalAttributeType2 val = "TooHeavy" />
        );
    );
    PostStates
    (
        PopulatedObjectClass "ConsequentActor"
        (
            <ObjectFrameClass ref = PersonObjectFrameClass />
            <DimensionSystem ref = RelativePosition />
            <Attribute ref = RelativeLocation expr = (a$+1) />
            <Attribute ref = RelativeTime expr = (t1$+1) />
            <Attribute ref = LiftingState val = "Lifting" />
        );
        PopulatedObjectClass "ConsequentActee"
        (
            <ObjectFrameClass ref = PersonObjectFrameClass />
            <PassiveParticipant val = "true" />
            <DimensionSystem ref = RelativePosition />
            <Attribute ref = RelativeLocation expr = (a$+1) />
            <Attribute ref = RelativeTime expr = (t1$+1) />
            <Attribute ref = PassiveIsLiftedState val = "Lifted" />
        );
    );
);
```

## 3. Ontology and KB for Schema: "Receiving/Delivering and Paying"

The behavior classes that are needed for this schema involve nested behaviors. Comprehendor's OntologyBuilder does not yet generate nested behaviors in behavior classes. However the following *object frame class* information items were auto-generated.

Natural language input:

A deliverable object is a common object.
A person can receive something.
A person can deliver something.
A person can pay a person.
A detective is a person.

The concept of a "deliverable" is used here as a high-level abstraction that includes any of: services, products, a report, etc. This is one of several possible ways to model the semantics of the input schema text "received the final report". (Although the behavior classes shown in the following section were hand-coded, not auto-generated, the following NL input could be used once this auto-generation feature is implemented in the Comprehendor NLU system).

If a person receives a deliverable then he/she does pay another person.
If a person delivers a deliverable then he/she is paid by another person.

### 3.1. Object Frame Classes

### 3.1.1. Lower Ontology: Auto-Generated Classes

```
ObjectFrameClass "PersonObjectFrameClass"
(
   <StructureTrait val = "Compound"/>

   AttributeTypes
   (
      AttributeType "PayingState"
      (
         <SuperType val = "Qualitative"/>

         <StateAttributeType val = "true" />

         "Values"
         (
            {
              "NotPaying",
              "Paying"
            }
         );
      );
```

```
        AttributeType "PassiveIsPayedState"
        (
           <SuperType val = "Qualitative"/>

           <StateAttributeType val = "true" />

           "Values"
           (
              {
               "NotPayed",
               "Payed"
              }
           );
        );

        AttributeType "ReceivingState"
        (
           <SuperType val = "Qualitative"/>

           <StateAttributeType val = "true" />

           "Values"
           (
              {
               "NotReceiving",
               "Receiving"
              }
           );
        );

        AttributeType "DeliveringState"
        (
           <SuperType val = "Qualitative"/>

           <StateAttributeType val = "true" />

           "Values"
           (
              {
               "NotDelivering",
               "Delivering"
              }
           );
        );
     );
  );

  ObjectFrameClass "DeliverableObjectObjectFrameClass"
  (
     <StructureTrait val = "Compound"/>

     DictionaryPriorWord
     (
        <DictionaryWordIsNoun val = "true" />

        English
        (
```

```
            { "deliverable",
              "deliverables" }
        );
    );

    Dictionary ( English
    (
        { "object",
          "objects" }
    ););

    HigherClasses ( { "CommonObjectFrameClass" } );
);


ObjectFrameClass "DeliverableObjectObjectFrameClass"
(
    <StructureTrait val = "Compound"/>

    AttributeTypes
    (
        AttributeType "PassiveIsReceivedState"
        (
            <SuperType val = "Qualitative"/>

            <StateAttributeType val = "true" />

            "Values"
            (
                {
                  "NotReceived",
                  "Received"
                }
            );
        );
    );
);


ObjectFrameClass "DeliverableObjectObjectFrameClass"
(
    <StructureTrait val = "Compound"/>

    AttributeTypes
    (
        AttributeType "PassiveIsDeliveredState"
        (
            <SuperType val = "Qualitative"/>

            <StateAttributeType val = "true" />

            "Values"
            (
                {
                  "NotDelivered",
                  "Delivered"
                }
            );
```

```
      );
   );
);

ObjectFrameClass "DetectiveObjectFrameClass"
(
   <StructureTrait val = "Compound"/>

   Dictionary ( English
   (
      { "detective",
        "detectives" }
   ););

   HigherClasses ( { "PersonObjectFrameClass" } );
);
```

## 3.2. Behavior Classes

### 3.2.1. Observations

The first two behavior classes shown here are for the actions “to deliver” and “to receive” (involving persons and deliverables).

Within the “paying” behavior classes, the nested behavior is in bold.

### 3.2.2. Listing

```
BehaviorClass "ReceiveBehaviorClass"
 (
   <BridgeObjectFrameClass ref = BehavioralStructuralParentClass />

   Dictionary ( English
   (
      {
        "receive",
        "received",
        "received",
        "receives",
        "receiving"
      }
   ););

   PriorStates
   (
      PopulatedObjectClass "AntecedentActor"
      (
         <ObjectFrameClass ref = PersonObjectFrameClass />
         <BinderSourceFlag val = "true" />
         <DimensionSystem ref = RelativePosition />
         <Attribute ref = RelativeLocation var = a$ />
         <Attribute ref = RelativeTime var = t1$ />
         <Attribute ref = ReceivingState val = "NotReceiving" />
```

```
        );
        PopulatedObjectClass "AntecedentActee"
        (
           <ObjectFrameClass ref = DeliverableObjectObjectFrameClass />
           <PassiveParticipant val = "true" />
           <DimensionSystem ref = RelativePosition />
           <Attribute ref = RelativeLocation expr = (a$+1) />
           <Attribute ref = RelativeTime expr = t1$ />
           <Attribute ref = PassiveIsReceivedState val = "NotReceived" />
        );
     );
     PostStates
     (
        PopulatedObjectClass "ConsequentActor"
        (
           <ObjectFrameClass ref = PersonObjectFrameClass />
           <DimensionSystem ref = RelativePosition />
           <Attribute ref = RelativeLocation expr = (a$+1) />
           <Attribute ref = RelativeTime expr = (t1$+1) />
           <Attribute ref = ReceivingState val = "Receiving" />
        );
        PopulatedObjectClass "ConsequentActee"
        (
           <ObjectFrameClass ref = DeliverableObjectObjectFrameClass />
           <PassiveParticipant val = "true" />
           <DimensionSystem ref = RelativePosition />
           <Attribute ref = RelativeLocation expr = (a$+1) />
           <Attribute ref = RelativeTime expr = (t1$+1) />
           <Attribute ref = PassiveIsReceivedState val = "Received" />
        );
     );
  );
  BehaviorClass "DeliverBehaviorClass"
   (
     <BridgeObjectFrameClass ref = BehavioralStructuralParentClass />

     Dictionary ( English
     (
        {
         "deliver",
         "delivered",
         "delivered",
         "delivers",
         "delivering"
        }
     ););

     PriorStates
     (
        PopulatedObjectClass "AntecedentActor"
        (
           <ObjectFrameClass ref = PersonObjectFrameClass />
           <BinderSourceFlag val = "true" />
           <DimensionSystem ref = RelativePosition />
           <Attribute ref = RelativeLocation var = a$ />
           <Attribute ref = RelativeTime var = t1$ />
           <Attribute ref = DeliveringState val = "NotDelivering" />
        );
```

```
    PopulatedObjectClass "AntecedentActee"
    (
      <ObjectFrameClass ref = DeliverableObjectObjectFrameClass />
      <PassiveParticipant val = "true" />
      <DimensionSystem ref = RelativePosition />
      <Attribute ref = RelativeLocation expr = (a$+1) />
      <Attribute ref = RelativeTime expr = t1$ />
      <Attribute ref = PassiveIsDeliveredState val = "NotDelivered" />
    );
  );
  PostStates
  (
    PopulatedObjectClass "ConsequentActor"
    (
      <ObjectFrameClass ref = PersonObjectFrameClass />
      <DimensionSystem ref = RelativePosition />
      <Attribute ref = RelativeLocation expr = (a$+1) />
      <Attribute ref = RelativeTime expr = (t1$+1) />
      <Attribute ref = DeliveringState val = "Delivering" />
    );

    PopulatedObjectClass "ConsequentActee"
    (
      <ObjectFrameClass ref = DeliverableObjectObjectFrameClass />
      <PassiveParticipant val = "true" />
      <DimensionSystem ref = RelativePosition />
      <Attribute ref = RelativeLocation expr = (a$+1) />
      <Attribute ref = RelativeTime expr = (t1$+1) />
      <Attribute ref = PassiveIsDeliveredState val = "Delivered" />
    );
  );
);

BehaviorClass "PayAfterReceivingBehaviorClass"
 (
  <CausalRule val = "true" />
  <BridgeObjectFrameClass ref = BehavioralStructuralParentClass />
  Dictionary ( English
  (
    { "pay", "payed", "paid", "pays", "paying" }
  ););
  PriorStates
  (
    PopulatedObjectClass "AntecedentActor"
    (
      <ObjectFrameClass ref = PersonObjectFrameClass />
      <BinderSourceFlag val = "true" />
      <DimensionSystem ref = RelativePosition />
      <Attribute ref = RelativeLocation var = a$ />
      <Attribute ref = RelativeTime var = t1$ />
      <Attribute ref = PayingState val = "NotPaying" />
      <Attribute ref = UniqueIdentityAttributeType var = q$ />  // (identity)
    );
    BehaviorClassReference
    (
      <BehaviorClass ref = ReceiveBehaviorClass />  //  -->> DEFINED-BEHAVIOR-CLASS
      <ParameterActor ref = PersonObjectFrameClass expr = q$ /> // (identity)
```

```
            <ParameterActee ref = DeliverableObjectObjectFrameClass />
            <ParameterExtra ref = PersonObjectFrameClass />
        );
        PopulatedObjectClass "AntecedentActee"
        (
            <ObjectFrameClass ref = PersonObjectFrameClass />
            <PassiveParticipant val = "true" />
            <DimensionSystem ref = RelativePosition />
            <Attribute ref = RelativeLocation expr = (a$+1) />
            <Attribute ref = RelativeTime expr = t1$ />
            <Attribute ref = PassiveIsPayedState val = "NotPayed" />
        );
    );
    PostStates
    (
        PopulatedObjectClass "ConsequentActor"
        (
            <ObjectFrameClass ref = PersonObjectFrameClass />
            <DimensionSystem ref = RelativePosition />
            <Attribute ref = RelativeLocation expr = (a$+1) />
            <Attribute ref = RelativeTime expr = (t1$+1) />
            <Attribute ref = PayingState val = "Paying" />
        );
        PopulatedObjectClass "ConsequentActee"
        (
            <ObjectFrameClass ref = PersonObjectFrameClass />
            <PassiveParticipant val = "true" />
            <DimensionSystem ref = RelativePosition />
            <Attribute ref = RelativeLocation expr = (a$+1) />
            <Attribute ref = RelativeTime expr = (t1$+1) />
            <Attribute ref = PassiveIsPayedState val = "Payed" />
        );
    );
);

BehaviorClass "PersonIsPaidAfterDeliveringBehaviorClass"
 (
    <CausalRule val = "true" />
    <BridgeObjectFrameClass ref = BehavioralStructuralParentClass />
    Dictionary ( English
    (
        { "pay", "payed", "paid", "pays", "paying" }
    ););
    PriorStates
    (
        PopulatedObjectClass "AntecedentActor"
        (
            <ObjectFrameClass ref = PersonObjectFrameClass />
            <BinderSourceFlag val = "true" />
            <DimensionSystem ref = RelativePosition />
            <Attribute ref = RelativeLocation var = a$ />
            <Attribute ref = RelativeTime var = t1$ />
            <Attribute ref = PayingState val = "NotPaying" />
        );
        BehaviorClassReference
        (
            <BehaviorClass ref = DeliverBehaviorClass />  //  -->> DEFINED-BEHAVIOR-CLASS
            <ParameterActor ref = PersonObjectFrameClass expr = q$ /> // (identity)
```

```
            <ParameterActee ref = DeliverableObjectObjectFrameClass />
            <ParameterExtra ref = PersonObjectFrameClass />
         );
         PopulatedObjectClass "AntecedentActee"
         (
            <ObjectFrameClass ref = PersonObjectFrameClass />
            <PassiveParticipant val = "true" />
            <DimensionSystem ref = RelativePosition />
            <Attribute ref = RelativeLocation expr = (a$+1) />
            <Attribute ref = RelativeTime expr = t1$ />
            <Attribute ref = PassiveIsPayedState val = "NotPayed" />
            <Attribute ref = UniqueIdentityAttributeType var = q$ />  // (identity)
         );
      );
      PostStates
      (
         PopulatedObjectClass "ConsequentActor"
         (
            <ObjectFrameClass ref = PersonObjectFrameClass />
            <DimensionSystem ref = RelativePosition />
            <Attribute ref = RelativeLocation expr = (a$+1) />
            <Attribute ref = RelativeTime expr = (t1$+1) />
            <Attribute ref = PayingState val = "Paying" />
         );
         PopulatedObjectClass "ConsequentActee"
         (
            <ObjectFrameClass ref = PersonObjectFrameClass />
            <PassiveParticipant val = "true" />
            <DimensionSystem ref = RelativePosition />
            <Attribute ref = RelativeLocation expr = (a$+1) />
            <Attribute ref = RelativeTime expr = (t1$+1) />
            <Attribute ref = PassiveIsPayedState val = "Payed" />
         );
      );
   );
```

## 4. Ontology and KB for Schema: “Councilmen and Demonstrators”

The ontology and KB for Winograd schema #1 are contained in Hofford (2014 (b)) “The ROSS User’s Guide and Reference Manual”.

## Appendix 3: ROSS Instance Models

These listings are of external instance models for selected schema sentences. The generated instance models contain *object instances*, not classes: the object instances refer back to the ontology classes from which they have been instantiated.

### 1. Instance Model for Schema: "Trophy and Suitcase" ("too big" variant)

The external instance model shown here is for the sentence: "The trophy did not fit in the suitcase because it was too *big*."

```
<?xml version="1.0" encoding="US-ASCII" standalone="yes"?>

<InstanceModel>

  <TranscriptHeader>
    <TextSource value="SubmittedFromWebClient">
    </TextSource>
  </TranscriptHeader>

  <ConceptualModel>

    <LocalContext contextId = "1">

      <MoodAndTense>
        Declarative-PastSimple
      </MoodAndTense>

      <StructuralParent name="EverydayObjectStructuralParentClass" >
        <Timeline name = "EverydayObjectStructuralParentClass.EverydayObjectDimensionSystem"/>
      </StructuralParent>

      <TimelineTimePoint value = "T01">
        <InstanceStructure>
          <Component>
            TrophyObjectFrameClass.TrophyObjectFrameClass-1 (trophy)
            <Attributes>
              <Attribute>
                EnclosableObjectFrameClass.FittingIntoState = FittingInto
              </Attribute>
              <Attribute>
                EnclosableObjectFrameClass.FunctionalAttributeType1 = TooBig
              </Attribute>
            </Attributes>
          </Component>
          <Component>
            SuitcaseObjectFrameClass.SuitcaseObjectFrameClass-1 (suitcase)
            <Attributes>
              <Attribute>
                ContainerObjectFrameClass.PassiveIsFittedIntoState = NotIsFittedInto
              </Attribute>
            </Attributes>
          </Component>
        </InstanceStructure>
      </TimelineTimePoint>
```

```
    <TimelineTimePoint value = "T02">
      <InstanceStructure>
        <Component>
           TrophyObjectFrameClass.TrophyObjectFrameClass-1 (trophy)
          <Attributes>
            <Attribute>
              EnclosableObjectFrameClass.PassiveIsFittedInsideContainerState = NotFittedInsideContainer
            </Attribute>
          </Attributes>
        </Component>
        <Component>
           SuitcaseObjectFrameClass.SuitcaseObjectFrameClass-1 (suitcase)
          <Attributes>
            <Attribute>
              ContainerObjectFrameClass.PassiveIsFittedIntoState = NotIsFittedInto
            </Attribute>
          </Attributes>
        </Component>
      </InstanceStructure>
    </TimelineTimePoint>

  </LocalContext>

 </ConceptualModel>

</InstanceModel>
```

End listing.

## 2. Instance Model for Schema: "Person Lifts Person" ("too weak" variant)

The external instance model for the person lifts person schema is as follows ("The man could not lift his son because he was too/so *weak*.").

```
<?xml version="1.0" encoding="US-ASCII" standalone="yes"?>

<InstanceModel>

  <TranscriptHeader>
    <TextSource value="DocumentFile">
    </TextSource>
    <DocumentFile name="Samples\Sentence-02.txt">
    </DocumentFile>
  </TranscriptHeader>

  <ConceptualModel>

    <LocalContext contextId = "1">

      <MoodAndTense>
         Declarative-PastSimple
      </MoodAndTense>

      <StructuralParent name="EverydayObjectStructuralParentClass" >
         <Timeline name = "EverydayObjectStructuralParentClass.EverydayObjectDimensionSystem"/>
      </StructuralParent>
```

```
<TimelineTimePoint value = "T01">
  <InstanceStructure>
    <Component>
       ManObjectFrameClass.ManObjectFrameClass-1 (man)
      <Attributes>
        <Attribute>
          PersonObjectFrameClass.LiftingState = NotLifting
        </Attribute>
         <Attribute>
          PersonObjectFrameClass.FunctionalAttributeType1 = TooWeak
        </Attribute>
      </Attributes>
    </Component>
    <Component>
       SonObjectFrameClass.SonObjectFrameClass-1 (son)
      <Attributes>
         <Attribute>
            PersonObjectFrameClass.PassiveIsLiftedState = NotLifted
         </Attribute>
      </Attributes>
    </Component>
  </InstanceStructure>
</TimelineTimePoint>

<TimelineTimePoint value = "T02">
  <InstanceStructure>
    <Component>
       ManObjectFrameClass.ManObjectFrameClass-1 (man)
      <Attributes>
         <Attribute>
            PersonObjectFrameClass.LiftingState = Lifting
         </Attribute>
      </Attributes>
    </Component>
    <Component>
       SonObjectFrameClass.SonObjectFrameClass-1 (son)
      <Attributes>
        <Attribute>
          PersonObjectFrameClass.PassiveIsLiftedState = NotLifted
        </Attribute>
      </Attributes>
    </Component>
  </InstanceStructure>
</TimelineTimePoint>

</LocalContext>

</ConceptualModel>

</InstanceModel>
```